\documentclass[lettersize,journal]{IEEEtran}

\usepackage{algorithm}
\usepackage{algpseudocode}
\usepackage{graphicx}
\usepackage{multirow}
\usepackage{subfigure}
\usepackage{ifpdf}
\usepackage{diagbox}
\usepackage{bbm}
\usepackage{amssymb}  
\usepackage{pifont}   
\usepackage{bbding}   
\usepackage{booktabs}
\usepackage{makecell} 
\usepackage{colortbl}  
\usepackage{xcolor}
\usepackage{float}
\usepackage{acronym}
\usepackage{amsmath}
\newcolumntype{I}{!{\vrule width 1pt}}  

\begin{document}
	
	\title{Text-Pass Filter: An Efficient Scene Text Detector}
	
	\author{Chuang~Yang,
		Haozhao~Ma,
		Xu~Han,
		Yuan~Yuan,~\IEEEmembership{Senior Member,~IEEE,}
		and~Qi~Wang,~\IEEEmembership{Senior Member,~IEEE}
		
		\thanks{
			Chuang~Yang, Haozhao~Ma, Yuan~Yuan, and Qi~Wang are with the School of Artificial Intelligence, OPtics and ElectroNics (iOPEN), Northwestern Polytechnical University, Xi'an 710072, P.R. China.
			
			Xu~Han is with the School of Computer Science, and with the School of Artificial Intelligence, OPtics and ElectroNics (iOPEN), Northwestern Polytechnical University, Xi'an 710072, Shaanxi, P. R. China. 
			
			E-mail: cyang113@mail.nwpu.edu.cn, haozhaoma@mail.nwpu.edu.cn, hxu04100@gmail.com, y.yuan.ieee@gmail.com, crabwq@gmail.com.}
		
		\thanks{Qi~Wang is the corresponding author.}
		
	}
	
	\markboth{~}%
	{Shell \MakeLowercase{\textit{et al.}}: A Sample Article Using IEEEtran.cls for IEEE Journals}
	
	
	\maketitle
	
	\begin{abstract}
		To pursue an efficient text assembling process, existing methods detect texts via the shrink-mask expansion strategy. However, the shrinking operation loses the visual features of text margins and confuses the foreground and background difference, which brings intrinsic limitations to recognize text features. We follow this issue and design Text-Pass Filter (TPF) for arbitrary-shaped text detection. It segments the whole text directly, which avoids the intrinsic limitations. It is noteworthy that different from previous whole text region-based methods, TPF can separate adhesive texts naturally without complex decoding or post-processing processes, which makes it possible for real-time text detection. Concretely, we find that the band-pass filter allows through components in a specified band of frequencies, called its passband but blocks components with frequencies above or below this band. It provides a natural idea for extracting whole texts separately. By simulating the band-pass filter, TPF constructs a unique feature-filter pair for each text. In the inference stage, every filter extracts the corresponding matched text by passing its pass-feature and blocking other features. Meanwhile, considering the large aspect ratio problem of ribbon-like texts makes it hard to recognize texts wholly, a Reinforcement Ensemble Unit (REU) is designed to enhance the feature consistency of the same text and to enlarge the filter's recognition field to help recognize whole texts. Furthermore, a Foreground Prior Unit (FPU) is introduced to encourage TPF to discriminate the difference between the foreground and background, which improves the feature-filter pair quality. Experiments demonstrate the effectiveness of REU and FPU while showing the TPF's superiority.
		
	\end{abstract}
	

	\section{Introduction}
	\label{Introduction}
	\IEEEPARstart{S}{cene} text understanding~\cite{xu2023morphtext,zhang2023arbitrary,wang2023region,9807447} is a hot topic in computer vision, which serves as a fundamental task in many practical applications (such as unmanned systems, bionic robots, and cognitive domain security defense). As an essential research branch of scene text understanding, scene text detection~\cite{DBLP:journals/tcsv/KeserwaniSLR22,DBLP:journals/tcsv/GuanGLTFWG22,DBLP:journals/tcsv/NandanwarSRLPAL22} is responsible for extracting the text regions from images. In this paper, we aim for real-time arbitrary-shaped text detection~\cite{DBLP:journals/tcsv/CaoZYZ22,DBLP:journals/tcsv/CaiLCDZWY21,DBLP:journals/tcsv/ChengCW20} from scenes.
	
	With the rapid development of deep learning and recent advances made in image segmentation~\cite{long2015fully,ronneberger2015u}, text detection~
	\cite{zhang2020opmp,wan2021self} achieves remarkable progress. Considering the various and complex shapes of scene texts, extracting the text regions from images via segmentation technology becomes a hot branch in the research of scene text detection. 
	
	Existing segmentation-based methods can be categorized into whole region-based approaches~\cite{10043020,9807447,zhu2021fourier,wang2020textray} and shrink-mask-based algorithms~\cite{yang2022zoom,DBLP:journals/tip/YangCXYW22,DBLP:journals/pami/LiaoZWYB23,9900473}. Whole region-based methods aim to segment entire text regions and directly extract the corresponding mask contours as final results. However, to separate overlapping texts, these methods often require complex decoding or post-processing steps to reconstruct text contours, deviating from their original design and introducing additional computational costs. In contrast, shrink-mask-based methods first roughly locate texts using shrink masks and then expand them to rebuild text contours, which streamlines the text assembly process. Nevertheless, the shrinking operation compromises the semantic integrity of the text and confuses the distinction between foreground and background, imposing intrinsic limitations on these methods' ability to recognize text features effectively.
	
	To remedy the problems that exist in the above two kinds of methods simultaneously, we propose an efficient framework in this work for accurate text detection with high inference speed. Specifically, inspired by the band-pass filter in electronics and signal processing, we find the operation mode that allows through components in a specified band of frequencies but blocks components with frequencies above or below this band is suitable for extracting every single text separately and simply. Based on this observation, the text-pass filter (TPF) is designed to extract texts from scenes.
	
	\begin{figure}
		\centering
		\includegraphics[width=.48\textwidth]{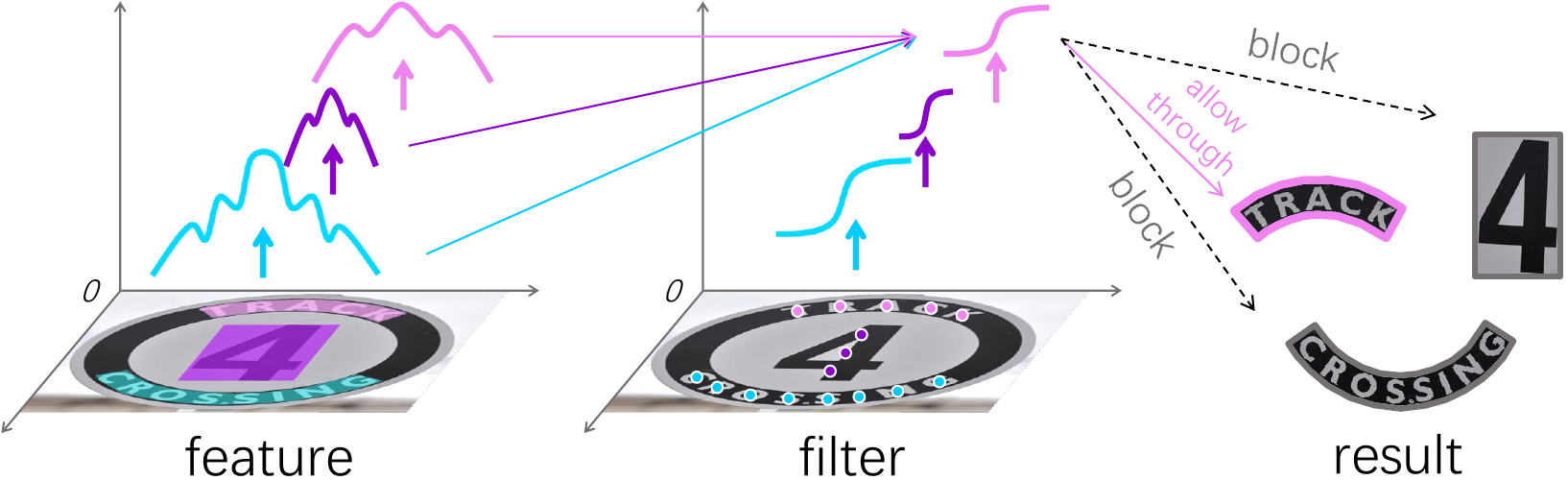}
		\caption{Motivation of the designed text-pass filter (TPF). It constructs a unique feature-filter pair for each text, where every feature represents a unique text. Each filter can only allow through the corresponding unique text feature and block the others, which helps detect every text separately in an efficient way.}
		\label{V1}
	\end{figure}
	
	TPF is constructed as an end-to-end convolutional neural network, which ensures straightforward training and inference processes. In the training stage, it encodes a unique feature-filter pair for every single text, where the feature and filter are treated as the passband and the band-pass filter, called pass-feature and text-pass filter, respectively. Different from shrink-mask-based methods, TPF focuses on whole text regions. It helps define a distinct semantic boundary between the text and the background, which enhances the recognition of text features. In the inference stage, TPF first combines all filters as a sieve and inputs the feature map that contains every text feature into it. All filters in the sieve then recognize their unique pass-feature in parallel to extract the corresponding text region. The efficient filtering process helps TPF separate adhesive texts without extra decoding or post-processing processes that exist in previous whole text region-based frameworks, which provides significant improvements in detection speed. 
	
	Meanwhile, different from normal objects, the ribbon-like text shape leads to the large aspect ratio problem. It further results in the feature inconsistency problem for the same text and the filter's limited recognition field problem, which makes filters are difficult to recognize the whole text region accurately. Considering the above problem, we design a Reinforcement Ensemble Unit (REU). It first measures the feature similarities of the same text and the feature differences with the others to enhance the feature consistency. Then, multiple filters of the same text are integrated to generate a strengthened filter according to feature similarities to enlarge the filter's recognition field to encourage the model to recognize whole text regions effectively. Furthermore, considering the feature-filter pair is extracted from the predicted text region, which makes the quality of the filter-feature pair deep relies on the model's ability to discriminate texts from the background, a Foreground Prior Unit (FPU) is introduced to encourage TPF to recognize the difference between the foreground and background. It helps the proposed efficient framework locate text instances more accurately. The contributions of this paper are summarized as follows:
	
	\begin{enumerate}
		\item Inspired by the band-pass filter (BPF) in electronics and signal processing, a text-pass filter (TPF) is proposed to formulate the text detection problem. It simulates the BPF operation mode to segment the whole region of each text directly and individually, which helps avoid the limitation of shrink-mask-based methods while making TPF enjoy a more efficient pipeline than existing whole region-based methods to separate adhesive texts.
		
		\item A Reinforcement Ensemble Unit (REU) is designed to measure the feature similarities of the same text and the feature differences with the others to enhance text feature consistency. Meanwhile, it integrates multiple filters of the same text according to feature similarities to generate a strengthened filter, which enlarges the filter's recognition field to encourage TPF to recognize whole text regions effectively.
		
		\item A Foreground Prior Unit (FPU) is introduced to discriminate the difference between the foreground and the background. It helps TPF to locate text instances more accurately, which encourages generating feature-filter pairs with high quality and suppressing false positives.
	\end{enumerate}
	
	The rest of the paper is organized as follows. Section~\ref{Related Work} introduces the previous works on text detection. Section~\ref{Methodology} describes the overall structure of TPF. The experimental results are discussed in Section~\ref{Experiments}. Section~\ref{Conclusion} concludes the paper.
	
	\section{Related Work}
	\label{Related Work}
	Deep learning-based text detection methods~\cite{zhang2022arbitrary,ye2023dptext,ye2023deepsolo++,baek2019character,9779460} have achieved significant progress recently, which can be divided into whole region-based methods and shrink-mask-based methods roughly. They will be introduced next.
	\subsection{Whole Text Region-based Methods}
	Researchers detect whole text regions and achieve superior performance initially~\cite{xu2020gliding} based on the detection framework~\cite{ren2015faster,liu2016ssd,redmon2016you,tian2019fcos}. Zhou~\textit{et al.}~\cite{zhou2017east} and He~\textit{et al.}~\cite{DBLP:conf/iccv/HeZYL17} followed the idea of Densebox to locate the text center and to regress the offsets between the center and four vertices for reconstructing text boxes. Different from the above methods, Liao~\textit{et al.}~\cite{DBLP:conf/aaai/LiaoSBWL17,liao2018textboxes++} proposed to regress the offsets between anchor vertices and text box vertices. To extract strong representative features of multi-oriented texts, Liao~\textit{et al.}~\cite{liao2018rotation} introduced active rotating filters to encode direction information of texts for enhancing rotation-invariant features. Recent research focus has shifted to more challenging irregular-shaped text detection. Some works~\cite{DBLP:conf/aaai/DengLLC18,xu2019textfield,liao2020mask} intuitively segment whole regions of texts to locate them directly based on segmentation methods~\cite{long2015fully,ronneberger2015u}. However, for separating texts that lie close to each other, Deng~\textit{et al.}~\cite{DBLP:conf/aaai/DengLLC18} and Xu~\textit{et al.}~\cite{xu2019textfield} proposed to encode the text direction information and assigned the pixels within text margins to the corresponding texts according to the information. Except for segmenting all pixels in whole text regions, many researches~\cite{wang2020textray,10043020,yang2022bip,9807447} design effective text representations to predict whole text contours. The works~\cite{DBLP:conf/ijcai/XueLZ19,zhang2019look,wang2020all} and Wang~\textit{et al.}~\cite{wang2020contournet} extracted a series of dense contour points via the regression and segmentation strategy to rebuild text contours, respectively. Zhu~\textit{et al.}~\cite{zhu2021fourier} were inspired by Fourier transformation and proposed to transform the text contour into compact signatures. Su~\textit{et al.}~\cite{9807447} encoded whole regions of texts into compact vectors via discrete cosine transformation. Although these methods have achieved superior performance for arbitrary-shaped scene text detection, they have to introduce the complex decoder or post-processing for separating adhesive texts.
	
	\begin{figure*}
		\centering
		\includegraphics[width=.85\textwidth]{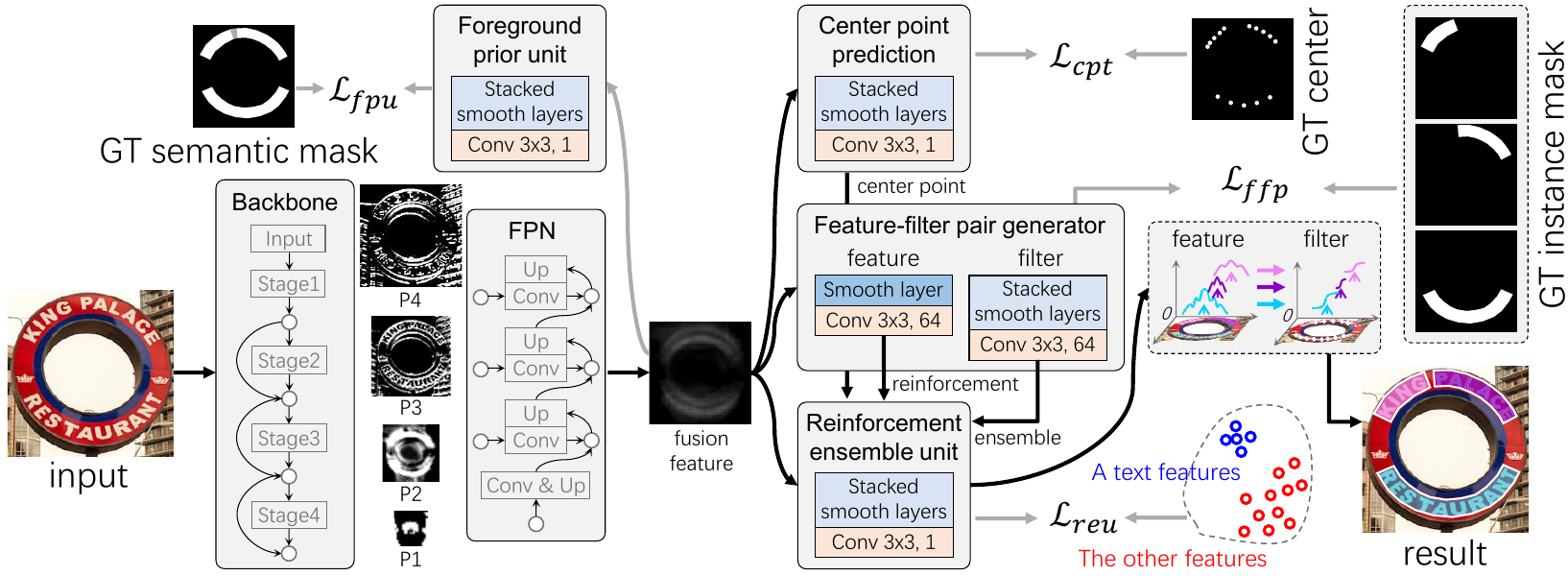}
		\caption{Overall architecture of the proposed TPF. It consists of a feature extractor, center point prediction header, feature-filter pair generator, Reinforcement Ensemble Unit (REU), and Foreground Prior Unit (FPU). The extractor includes the backbone and FPN and the corresponding output is a concatenated feature map with the size of $\frac{H}{4}$,$\frac{W}{4}$. The center point prediction header is responsible for locating text instances. REU is designed for encouraging the feature-filter pair generator to produce feature-filter pair with high quality for each text. FPU is introduced to help recognize text center points more accurately. The black flow and gray flow illustrate the forward and backward propagation of the whole training process. Particularly, the gray arrows are the inference-only operators. They bring no extra computational cost for the testing process.}
		\label{V2}
	\end{figure*}
	
	\subsection{Shrink-Mask-based Methods}
	To simplify the detection pipeline and speed up the inference speed for pursuing real-time requirements of practical applications, researchers design a shrink-mask expansion strategy to extract texts. Wang~\textit{et al.}~\cite{wang2019shape} first predicted multiple shrink-masks with different shrunk scales and whole regions via segmentation methods directly. All pixels within larger shrink-masks then were assigned to smaller ones step by step until whole regions become the aforementioned larger ones. Considering the low efficiency of the stepwise expansion process, Wang~\textit{et al.}~\cite{wang2019efficient,DBLP:journals/pami/WangXLLLYLS22} predicted shrink-masks with one specific scale, and whole regions then were clustered into shrink-masks according to the idea of the clustering algorithm. Different from the above methods that extra predicted whole regions, Liao~\textit{et al.}~\cite{liao2020real,DBLP:journals/pami/LiaoZWYB23} could reconstruct text contours according to shrink-masks merely. They computed an expansion offset via the area and perimeter of shrink-masks and expanded shrink-mask contours by the offset to rebuild text contours. Yang~\textit{et al.}~\cite{DBLP:journals/tip/YangCXYW22} observed the strategy that generating shrink-masks via the area and perimeter fail to represent hourglass texts. The authors presented to shrink the text masks according to the text shapes for enhancing the model's fitting ability effectively. Yang~\textit{et al.}~\cite{9900473} found the detection results of the method~\cite{liao2020real} rely on the accuracy of predicted shrink-masks deeply, which leads to sensitive rebuilt text contours. They proposed to encode the shrink-mask and expansion offset separately for pursuing robust reconstructed results. Different from the aforementioned methods, Yang~\textit{et al.}~\cite{yang2022zoom} constructed an efficient encoder to further improve the detection speed while ensuring the strong representativity of the extracted features. Though the shrink-mask expansion strategy helps separate adhesive texts efficiently, the broken semantic integrity and the feature confusion between the foreground and background bring intrinsic limitations for these methods to recognize text features accurately.
	
	\section{Methodology}
	\label{Methodology}
	This section first introduces the overall architecture of the network that is constructed based on the designed text-pass filter (TPF). The structure and operation detail of Reinforcement Ensemble Unit (REU) then is illustrated. Next, Foreground Prior Unit (FPU) is described. In the end, the loss function used for supervising whole network is formulated.
	
	\begin{figure}
		\centering
		\includegraphics[width=.45\textwidth]{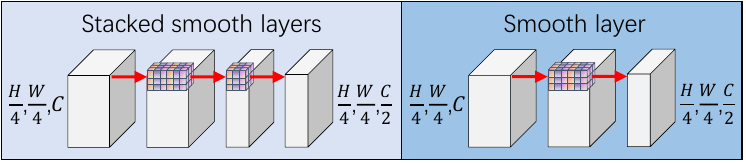}
		\caption{Structure details of stacked smooth layers and smooth layer. They are composed of 3$\times$3 convolution al layers mainly.}
		\label{V3}
	\end{figure}
	
	\begin{figure*}
		\centering
		\includegraphics[width=.8\textwidth]{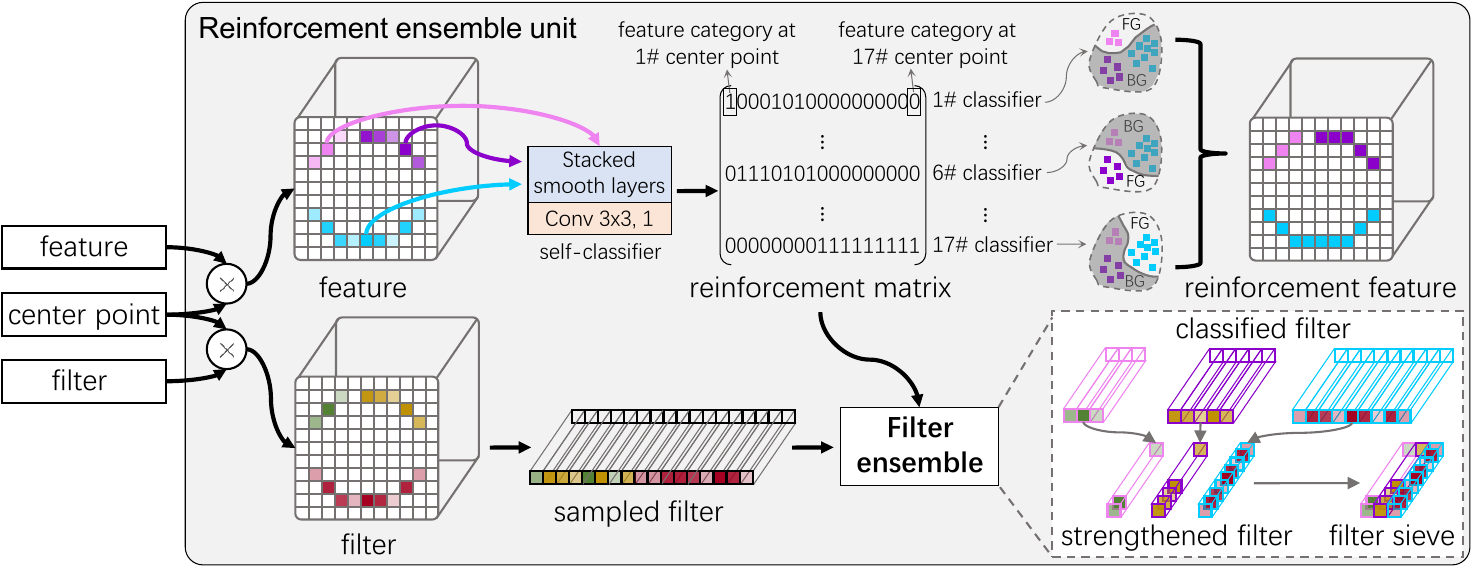}
		\caption{Structure details of the designed REU. The feature and filter are the outputted feature map and filter map from the feature-filter pair generator. The center point denotes the point coordinates predicted by the center point prediction header. The text features and filters first are sampled from the feature map and filter map according to center point coordinates. The feature consistency and filter recognition ability then are enhanced and strengthened under the guidance of the reinforcement matrix, respectively. In the end, the strengthened filters are combined as a filter sieve by the filter ensemble algorithm (as illustrated in Algorithm~\ref{algorithm1}) for detecting texts efficiently.}
		\label{V4}
	\end{figure*}
	
	\subsection{Overall Architecture}
	\label{Overall Architecture}
	The overall architecture of the proposed TPF is illustrated in Fig.~\ref{V2}. Following the design of traditional feature extractor~\cite{he2016deep,lin2017feature}, TPF first extracts strong representative features by the combination of backbone and feature pyramid network (FPN). It adopts ResNet as the backbone to generate multiple-sized feature maps (including the $\frac{1}{4},\frac{1}{8},\frac{1}{16},\frac{1}{32}$ input image sizes), where smaller feature maps help our model distinguish texts from the background and the larger ones encourage TPF focus on text details. Different from normal objects, considering the large aspect ratio of the text instance, the FPN is constructed behind the backbone to extract a fusion feature with a $\frac{1}{4}$ image size for detecting texts, which helps improve the recall performance of detection results. 
	
	The fusion feature is then fed into the center point prediction header, feature-filter pair generator, and reinforcement ensemble unit (REU) for segmenting text masks. Specifically, for the center point prediction header, it consists of stacked smooth layers  and one convolution layer with a 1$\times$1 kernel. The stacked smooth layers (the detail can be found in Fig.~\ref{V3}) are composed of two 3$\times$3 convolution layers with $C$ and $\frac{C}{2}$ channels respectively, which are used for smoothing the gap between fusion features and the final output center point map. Inputting the fusion feature into the prediction header, it is smoothed from $C$ channels to $\frac{C}{2}$ channels and a $\frac{1}{4}$ image-sized center point map is segmented for locating texts. The feature-filter pair generator includes two branches for generating a text feature map and constructing the corresponding filter map, respectively. The two branches enjoy a similar structure with the center point prediction header except that the predicted maps are with 64 channels. By combining the feature and filter maps and the predicted center point locations, the text features and filters can be extracted. 
	
	Normally, instance masks can be obtained separately by simulating the band-pass filter operation mode that the filters allow through their unique pass-feature and block other features. However, the feature difference problem and the filter local-sensitive problem brought by the large aspect ratio make it hard to segment whole text masks directly. Considering the above problems, the REU takes the sampled text features and filters as input to enhance the feature consistency and filter reliability for encouraging the model to recognize whole text regions (details are described in Section~\ref{Reinforcement Ensemble Unit} and Fig.~\ref{V3}). The processed text filters are combined as a filter sieve, where all filters allow through their unique pass-features to extract all text masks in parallel.
	
	\subsection{Reinforcement Ensemble Unit}
	\label{Reinforcement Ensemble Unit}
	Considering the large aspect ratio problem of the text results in the feature inconsistency problem and the filter's limited recognition field problem for the same text, which makes it hard for filters to recognize the whole text features effectively, REU is designed for enhancing feature consistency of the same text while enlarging the filter's recognition field. 
	
	\begin{algorithm}  
		\caption{Filter Ensemble}  
		\begin{algorithmic}[1]
			\Require The sampled filters $\mathbf{F}_{fi}$ and reinforcement matrix $\mathbf{M}$;
			\Ensure The ensemble filter sieve $\mathbf{F}_{fis}$;
			\State $\mathbf{F}_{fis} \gets [~]$
			\State $v \gets$ sum($\mathbf{M}$, dim=0)
			\State $v_{sort}\gets$ sort($v$, ascending=False)
			\For{$u$ in unique($v_{sort}$)}
			\State $idx \gets$ argwhere($v$==$u$)
			\State $\mathbf{M}_u \gets \mathbf{M}[\mathbf{M}[:,idx]]$
			\State$\mathbf{M}_c \gets \mathbf{M}_u[0]$
			\For{$r$ in range(1,$\mathbf{M}_u$.shape[0])}
			\State$\mathbf{M}_c \gets \mathbf{M}_c \times \mathbf{M}_u[r]$
			\EndFor
			\State $f_c \gets \mathbf{F}_{fi}[$argwhere($\mathbf{M}_c$==1)$]$~~//$f_c$ is classified filter
			\State $f_{str} \gets $mean($f_c$)~~//$f_{str}$ denotes strengthened filter
			\State $\mathbf{F}_{fis} \gets f_{str}$
			\EndFor
			\State $\mathbf{F}_{fis} \gets$ concate($\mathbf{F}_{fis}$, axis=1)~~//$\mathbf{F}_{fis}$ is filter sieve
		\end{algorithmic}  
		\label{algorithm1}
	\end{algorithm}
	
	For \textbf{enhancing feature consistency} of the same text, as shown in Fig.~\ref{V4}, REU takes center point coordinates and feature map as input. It first extracts sampled text features $\mathbf{F}_{fe}\in \mathbb{R} ^{n\times64}$ from the feature map according to the $n$ center point coordinates, where each feature vector ${f}_{fe}\in \mathbb{R} ^{1\times64}$ represents a part of text features. 
	
	REU then constructs a self-classifier for assigning $n$ feature vectors to different text instances, where the self-classifier is implemented as a combination of a stacked smoothing layer and a convolutional layer featuring a 3$\times$3 kernel with single-channel output, and the number of self-classifier output changed dynamically according to the predicted text mask. For example, given $n$ center points, the classifier sequentially treats each point as a reference and predicts whether all other points belong to the same text instance as this reference (outputting 0 or 1 for each prediction), ultimately generating a rank $n$ reinforcement matrix. The output dimensions of this process are entirely determined by the number of center points. Concretely, in the inference stage, it predicts a reinforcement matrix $\mathbf{M}\in \mathbb{R} ^{n\times n}$, where each row of M represents the binary classification results of the self-classifier to all ${f}_{fe}$. For $r$th row of M, the $c$th column binary value represents that the self-classifier assigns $c$th ${f}_{fe}$ and $r$th ${f}_{fe}$ into one text instance if the corresponding value is 1. The self-classifier operation forces the feature vectors that belong to the same text instance to enjoy strong similarities with each other. Meanwhile, it strengthens the feature vector differences when they belong to different texts. The above advantages of RUE help filters avoid segmenting the same text as multiple ones and improve the precision performance of detection results effectively.
		
	\begin{figure*}
		\centering
		\includegraphics[width=.9\textwidth]{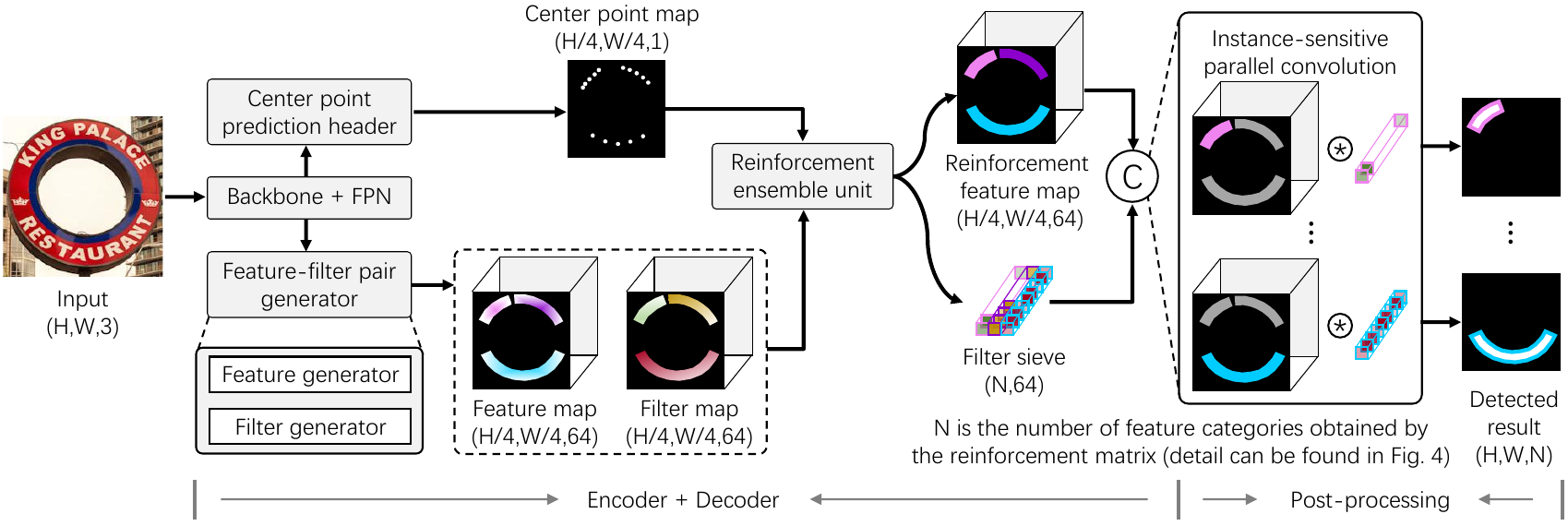}
		\caption{Visualization of inference process. Structure details of center point prediction header, feature-filter pair generator, REU, and FPU can be found in Fig.~\ref{V2}, Fig.~\ref{V3}, and Fig.~\ref{V4}.}
		\label{V41}
	\end{figure*}
	
	For \textbf{enlarging filter's recognition field}, REU takes the same sampling process as the text features. It first extracts sampled text filters $\mathbf{F}_{fi}\in \mathbb{R} ^{n\times64}$ by the combination of $n$ center point coordinates and filter map, where $\mathbf{F}_{fi}$ consists of $n$ filters ${f}_{fi}\in \mathbb{R} ^{1\times64}$. All filters are then strengthened and combined as a filter sieve $\mathbf{F}_{fis}\in \mathbb{R} ^{m\times64}$ for segmenting all whole text masks in parallel, where $m$ denotes the number of text instances of the input image. Specifically, as illustrated in Algorithm~\ref{algorithm1}, given the sampled filters $\mathbf{F}_{fi}$ and the reinforcement matrix $\mathbf{M}$ that is predicted according to the sampled text features $\mathbf{F}_{fe}$, the filter ensemble algorithm generates the filter sieve $\mathbf{F}_{fis}$ through the following four steps mainly: (1) performing an add operation along the column direction of the matrix $\mathbf{M}$ to obtain a filter importance sequence $v\in \mathbb{R} ^{n\times1}$; (2) sorting the sequence $v$ in descending order to generate a priority sequence $v_{sort}\in \mathbb{R} ^{n\times1}$; (3) determining the maximum value corresponding column index $idx_{max}$ in $v$; (4) extracting the rows from $\mathbf{M}$ according to the index $idx_{max}$; (5) intersecting those rows from M to obtain a strengthened row and extracting each location on the row corresponding filter for collecting all filters that belong to the same text instance at first. Then, merging those filters via the mean value method to generate the final strengthened filter $f_{str}$, which is responsible for filtering the matched text; (6) repeating (1)--(5) steps to produce $m$ strengthened filters and combined them as a filter sieve $\mathbf{F}_{fis}$. The strengthened filter helps enlarge the recognition field of the sampled filter and the filter sieve accelerates the inference speed significantly.
	
	\subsection{Foreground Prior Unit}
	\label{Foreground Prior Unit}
	As illustrated in Fig.~\ref{V2} and Fig.~\ref{V4}, TPF extracts feature-filter pair from the feature and filter maps according to the predicted center point coordinates, which leads to the feature-filter pair quality deep relies on the model's foreground recognition ability. To help TPF to locate text instances accurately, FPU is introduced to encourage foreground discrimination from the background under the guidance of prior semantic information.
	
	Considering the center point prediction header, feature-filter pair generator, and REU rely on the fusion feature, FPU takes it as input to strengthen the corresponding semantic feature. FPU first smooths the fusion feature via a shallow smooth layer, which helps fuse high-level and low-level features while ensuring an effective backward propagation of semantic information to the fusion feature. FPU then generates a binary mask through a 3$\times$3 convolutional layer to learn the feature difference between the foreground and background by supervising the predicted mask with GT semantic mask.
	
	\subsection{Inference Process}
	\label{Inference Process}
	We have introduced the overall architecture of TPF in Section~\ref{Overall Architecture} and the details of the proposed REU and FPU in Section~\ref{Reinforcement Ensemble Unit} and \ref{Foreground Prior Unit}. To show the efficiency of our method, the inference pipeline is illustrated in this section.
	
	Concretely, it is shown in Fig.~\ref{V41}, given an input image, the binary mask map of center point, text feature map, and text filter map can be obtained from center point prediction header and feature filter pair generator at first. To remedy the scene text large aspect ratio problem, the three maps then are fed into REU to enhance the consistency of feature map and enlarge the recognition field of filter map according to reinforcement matrix $\mathbf{M}$ (the process can be referred in Section~\ref{Reinforcement Ensemble Unit}). Following the motivation of TPF, the reinforcement feature map and filter sieve generated from REU are treated as the text pass-feature and text pass-filter, respectively. In the end, every whole region of text instance can be obtained via the instance-sensitive parallel convolution-based post-processing. Specifically, since each text pass-filter is sensitive to the corresponding text pass-feature only, all filters can recognize their pass-features in parallel to detect all texts separately without the interference of the adhesion problem. It ensures efficient post-processing and makes TPF runs faster than existing whole region-based text detection methods.
	
	\begin{figure*}
		\centering
		\includegraphics[width=.85\textwidth]{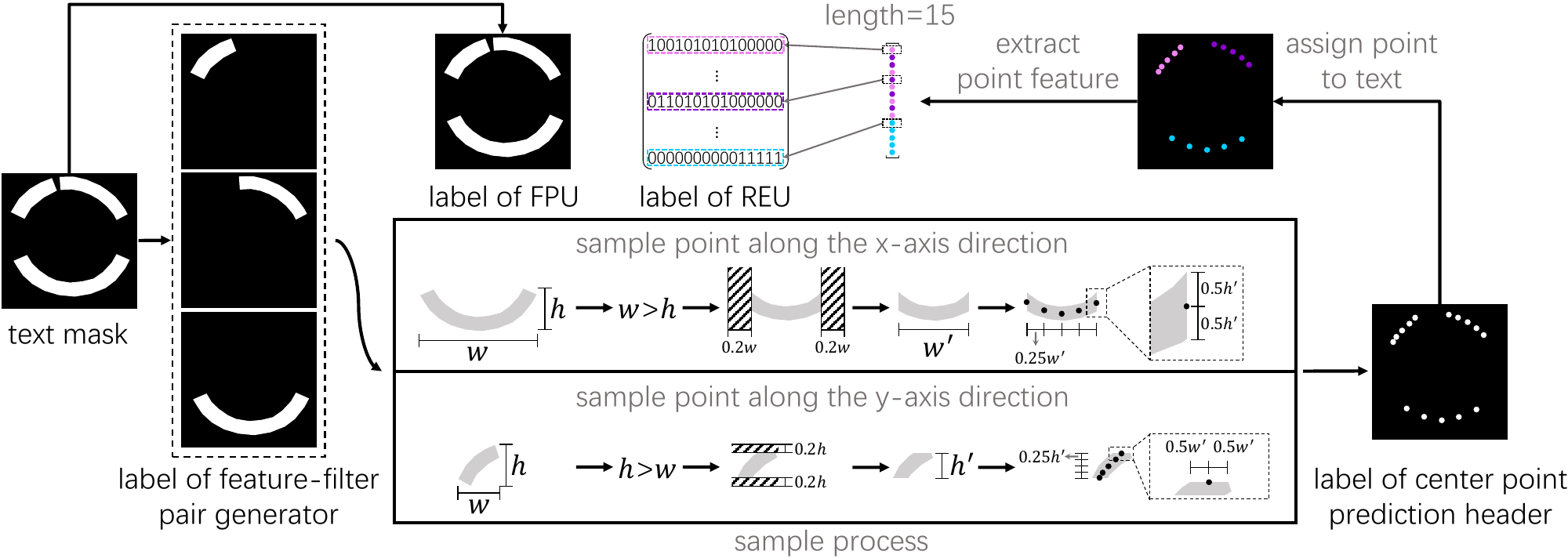}
		\caption{Visualization of label generation process. For the four branches of center point prediction header, feature-filter pair generator, REU, and FPU, the corresponding label are shown in detail.}
		\label{V5}
	\end{figure*}
	
	\subsection{Label Generation}
	\label{Label Generation}
	As illustrated in Fig.~\ref{V2}, the proposed network includes four prediction branches (FPU, center point prediction header, feature-filter pair generator, and REU). This section describes the corresponding label generation process in detail.
	
	For \textbf{feature-filter pair generator}, it is designed for generating feature-filter pair for each text instance. To effectively supervise this branch in the training process, we extract all text instance masks from a binary mask one by one and combines them to obtain the corresponding label $\mathbf{F}_{ins}\in \mathbb{R} ^{\frac{H}{4}\times\frac{W}{4}\times m}$, where $m$ denotes the number of text instances of the input image. $H$ and $W$ are the height and width, respectively.
	
	For \textbf{center point prediction header}, this branch is responsible for locating texts. The outputted center point coordinates are used for obtaining the feature-filter pair of the text instance from the predicted feature and filter maps from the feature-filter pair generator in the inference process. Given a $\mathbf{F}_{ins}$, we sample center points from every text instance mask by the sampling process in Fig.~\ref{V5}. Concretely, for a specific text instance, the process first computes its height $h$ and width $w$ along the y-axis and x-axis directions simultaneously. Then, the x-axis direction is determined as the sampling direction if $w>h$ else performing the sampling process along the y-axis direction. Next, the middle part of the text instance is extracted as the valid sampling area. In the end, $n$ center points are sampled from the valid area along the determined axis direction through equidistant sampling process. The sampled points of all texts are integrated into one mask for generating the final center point label.
	
	For \textbf{Reinforcement Ensemble Unit}, we introduce this structure for enhancing the feature consistency of the same text while strengthening the filter's recognition ability (as illustrated in Section~\ref{Reinforcement Ensemble Unit} and Fig.~\ref{V4}). The label of this branch can be obtained by combining the labels of feature-filter pair generator and center point prediction header. Specifically, all center points are first assigned to different text instances according to $\mathbf{F}_{ins}$, and then a matrix label $\mathbf{M}\in \mathbb{R} ^{n\times n}$ is generated based on the point coordinates and point assignment situations. $\mathbf{M}$ is a binary matrix, $r$ th row of $\mathbf{M}$ represents the classification results of $r$ th self-classifier to all points and $c$ th column of $\mathbf{M}$ denotes the classification results of all self-classifiers to $c$ th points ($0<r<n,0<c<n$). For $r$ th row of $\mathbf{M}$, the value of $c$ column being `1' if the $c$ th point and $r$ th point belong to the same text instance else `0'. 
	
	For \textbf{Foreground Prior Unit}, it is proposed to guide our method to distinguish text instances from the background more accurately. The text binary mask is adopted as the label of FPU. As shown in Fig.~\ref{V5}, all texts are drawn in the mask. The foreground and background are labeled as `1' and `0'. 
	
	\begin{figure*}
		\centering
		\subfigure[Instance geometry characteristics analysis of training dataset]{
			\begin{minipage}[b]{0.98\linewidth}
				\includegraphics[width=1\linewidth]{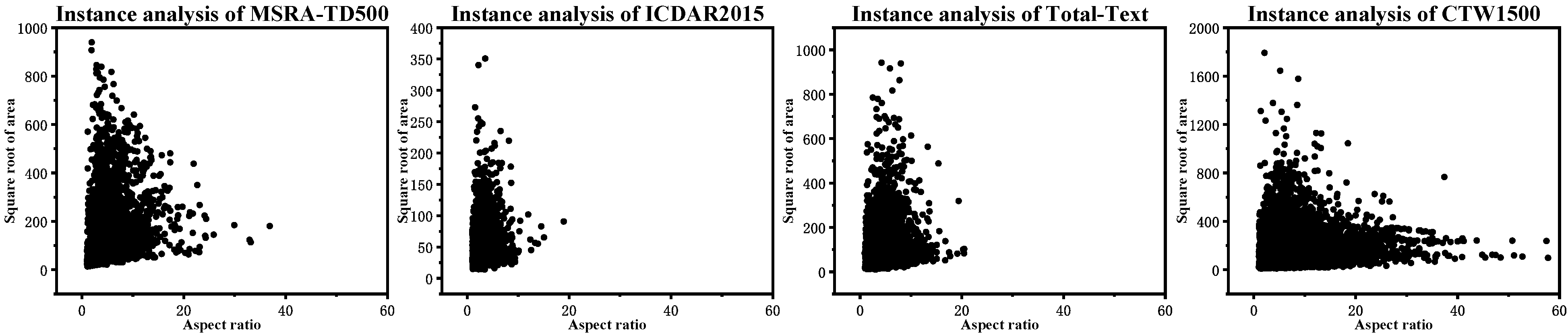}
		\end{minipage}}
		\subfigure[Instance geometry characteristics analysis of testing dataset]{
			\begin{minipage}[b]{0.98\linewidth}
				\includegraphics[width=1\linewidth]{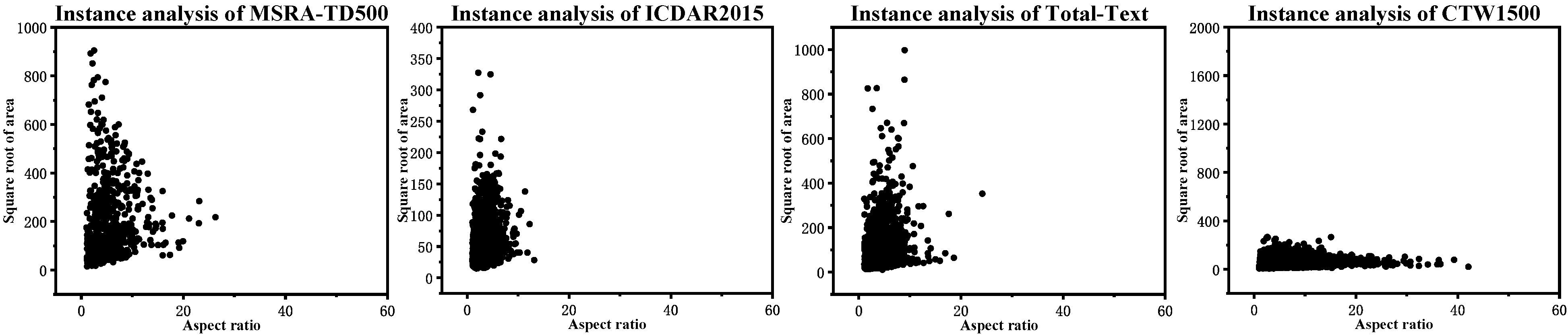}
		\end{minipage}}
		\caption{Analysis of geometry characteristics of text instances on multiple public datasets (MSRA-TD500, CTW1500, Total-Text, and ICDAR2015). (a) and (b) are the training and testing dataset, respectively.}
		\label{V6}
	\end{figure*}
	
	\subsection{Loss Function}
	\label{Loss Function}
	The proposed TPF generates feature-filter pair for each text instance in the training process and extracts all text masks separately by the pass-feature recognition of the filter in the inference process. To predict the feature-filter pair efficiently, FPU is introduced to help our model recognize text instances, which encourages TPF to locate text center points more accurately. Meanwhile, REU is constructed to enhance the filter recognition ability and feature consistency, which improves the quality of the text mask generated by the feature-filter pair while simplifying post-processing. 
	
	For supervising the four branches that exist in TPF effectively, we formulate the loss function ${\cal L}$ as follows:
	\begin{eqnarray}
		\label{e1}
		{\cal L}=\alpha {\cal L}_{fpu}+\beta {\cal L}_{cpt}+\mu {\cal L}_{ffp}+\lambda {\cal L}_{reu} ,
	\end{eqnarray}
	where ${\cal L}_{fpu}$, ${\cal L}_{cpt}$, ${\cal L}_{ffp}$, and ${\cal L}_{reu}$ are loss functions for supervising FPU, center point prediction header, feature-filter pair generator, and REU, respectively. $\alpha$, $\beta$, $\mu$, and $\lambda$ are the corresponding importance weights of them.
	
	For ${\cal L}_{cpt}$ and ${\cal L}_{reu}$, they are responsible for evaluating the accuracy of the predicted center points and reinforcement matrix. As illustrated in Section~\ref{Label Generation} and Fig.~\ref{V5}, the label of the center point and reinforcement matrix enjoy the characteristics of sparsity, discontinuity, and imbalance between positive and negative samples. Cross entropy (CE) loss is proposed for binary classification tasks initially, but it performs badly when handling the sample imbalance. Considering the deficiency of CE loss, focal loss ${\cal L}_{fl}$~\cite{lin2017focal} is designed based on CE loss. It forces the model to focus on the positive samples by reducing the weights of negative samples, which is suitable for supervising the center point prediction header and REU:
	\begin{eqnarray}
		\label{e2}
		\begin{gathered}
			{\cal L}_{fl}(p_t)=-(1-p_t)^\gamma  log(p_t),\\
			p_t=\begin{cases}
				p, & \text{ if } y=1 \\
				1-p, & \text{ otherwise}\end{cases}.
		\end{gathered}
	\end{eqnarray}
	
	Considering the superiority of ${\cal L}_{fl}$, we formulate ${\cal L}_{cpt}$ and ${\cal L}_{reu}$ through it as follows:
	\begin{eqnarray}
		\label{e3}
		\begin{gathered}
			{\cal L}_{cpt}={\cal L}_{fl}(p_{cpt}),\\
			{\cal L}_{reu}={\cal L}_{fl}(p_{reu}),
		\end{gathered}
	\end{eqnarray}
	where $p_{cpt}$ and $p_{reu}$ are the predicted probability values of the center point and reinforcement matrix element.
	
	Since the label of FPU and feature-filter pair generator are continuous, we construct the corresponding loss function ${\cal L}_{fpu}$ and ${\cal L}_{ffp}$ by dice loss~\cite{milletari2016v}:
	\begin{eqnarray}
		\label{e4}
		\begin{gathered}
			{\cal L}_{dl}(\mathbf{P},\mathbf{\widehat{P}})= 1-\frac{2\times| {\mathbf{P}}\cap {\mathbf{\widehat{P}}}|+\varepsilon}{| {\mathbf{P}} |+| {\mathbf{\widehat{P}}}|+\varepsilon},\\
			{\cal L}_{fpu}={\cal L}_{dl}(\mathbf{P}_{fpu},\mathbf{\widehat{P}}_{fpu}),\\
			{\cal L}_{ffp}={\cal L}_{dl}(\mathbf{P}_{ffp},\mathbf{\widehat{P}}_{ffp}),
		\end{gathered}
	\end{eqnarray}
	where $\mathbf{P}_{fpu}$ and $\mathbf{P}_{ffp}$ are the predicted foreground mask and reinforcement matrix. $\mathbf{\widehat{P}}_{fpu}$ and $\mathbf{\widehat{P}}_{ffp}$ are the corresponding label (generation process can be referred in Fig.~\ref{V5}). 
	
	The center point prediction header, feature-filter pair generator, and REU participate in the process of segmenting text masks directly. Different from them, FPU is introduced as an assistant module to help the above three branches recognize texts more accurately. Considering the different importance of the four branches, we set $\alpha$, $\beta$, $\mu$, and $\lambda$ as 0.7, 1, 1, and 1, respectively in the following experiments of this paper.
	
	\section{Experiments}
	\label{Experiments}
	TPF is designed for detecting arbitrary-shaped scene texts efficiently. In this section, the effectiveness of the proposed approach and its enhancement modules first are verified in the ablation study. The superior comprehensive performance of TPF then is shown by comparing it with existing methods.
	
	\subsection{Datasets}
	\textbf{SynthText}~\cite{gupta2016synthetic} provides 800 thousand scene text images and 8 million text instances to help pre-train detection networks for improving the model's generalization ability, where each image is synthesized manually by placing multiple randomly generated texts in a natural scene image.
	
	\textbf{MSRA-TD500}~\cite{yao2012detecting} consists of line-level multi-oriented texts, which include English and Chinese simultaneously. It has 300 training and 200 testing images that are sampled from indoors. Considering the fewer training samples, we follow previous works to introduce 400 images from HUST-TR400~\cite{yao2014unified} to build the training dataset.
	
	\textbf{ICDAR2015}~\cite{karatzas2015icdar} is proposed in the Incidental Scene Text 2015 challenge of the Robust Reading Competition. It assigns 1000 images and 500 images for training and testing the model, respectively. The multi-scaled text instances and the complex image background bring challenges for text detection.
	
	\textbf{Total-Text}~\cite{ch2017total} includes 1255 training images and 300 testing images. Different from the line-live text instances, the word-level texts in this dataset can demonstrate the model's superior ability to separate adhesive texts effectively.
	
	\textbf{CTW1500} is constructed by 1500 scene images, where 1000 images are used for training models and 500 images are responsible for testing models. The line-level curved text instances of this dataset can evaluate the model's ability for detecting irregular-shaped scene texts integrally.
	
	We will demonstrate the superiority of our TPF on the above public scene text detection datasets next. For well analyzing the model performance to texts with different scales, aspect ratios, and shapes, we visualize the text instance information in Fig.~\ref{V6}. It can be found that the text instances of the training and testing data enjoy a similar scale and aspect ratio distribution for MSRA-TD500, Total-Text, and ICDAR2015 datasets, which is beneficial for evaluating model performance without interference. For the CTW1500 dataset, the training sample scale is larger than the testing sample scale a lot. The performance of the model that is trained and tested on samples with different scales can be explored on this dataset. Moreover, the line-level datasets (MSRA-TD500 and CTW1500) enjoy a larger aspect ratio than word-level datasets (Total-Text and ICDAR2015). The model's ability for dealing with samples with different aspect ratios can be analyzed by the cross-evaluation experiments on these datasets.
	
	\begin{table*}[]
		\renewcommand{\arraystretch}{1}
		\setlength{\tabcolsep}{2.mm}
		\caption{Performance analysis of the models with different settings on the MSRA-TD500 benchmark. The ``short side: 736'' denotes the testing image is resized to the 736 proportionally along its short side. The model is not pre-trained on any public datasets.}
		\centering
		\small
		\begin{tabular}{clcccccc}
			\toprule
			\multicolumn{8}{c}{Image size for testing (short side: 736)} \\ 
			\midrule
			\multicolumn{1}{c}{\#} & \multicolumn{1}{l}{Methods}   & \multicolumn{1}{c}{REU} & \multicolumn{1}{c}{FPU} & \multicolumn{1}{c}{Precision(\%) $\uparrow$} & \multicolumn{1}{c}{Recall(\%) $\uparrow$}   & \multicolumn{1}{c}{F-measure(\%) $\uparrow$} & \multicolumn{1}{c}{FPS$\uparrow$}   \\ 
			\midrule
			\multicolumn{1}{c}{1}  & \multicolumn{1}{l}{baseline}  & \multicolumn{1}{c}{\ding{55}}    & \multicolumn{1}{c}{\ding{55}}       & \multicolumn{1}{c}{87.9}  & \multicolumn{1}{c}{79.2} & \multicolumn{1}{c}{83.3}               & \multicolumn{1}{c}{33.6}      \\ 
			\multicolumn{1}{c}{2}  & \multicolumn{1}{l}{baseline+} & \multicolumn{1}{c}{\ding{52}}   & \multicolumn{1}{c}{\ding{55}}       & \multicolumn{1}{c}{89.7~(\textbf{{1.8$\uparrow$})}}  & \multicolumn{1}{c}{80.7~(\textbf{{1.5$\uparrow$})}} & \multicolumn{1}{c}{85.0~(\textbf{{1.7$\uparrow$})}}              & \multicolumn{1}{c}{36.2~(\textbf{{2.6$\uparrow$})}}      \\ 
			\multicolumn{1}{c}{3}  & \multicolumn{1}{l}{baseline+} & \multicolumn{1}{c}{\ding{52}}   & \multicolumn{1}{c}{\ding{52}}      & \multicolumn{1}{c}{89.9~(\textbf{{0.2$\uparrow$})}}  & \multicolumn{1}{c}{82.8~(\textbf{{2.1$\uparrow$})}} & \multicolumn{1}{c}{86.2~(\textbf{{1.2$\uparrow$})}}              & \multicolumn{1}{c}{37.7~(\textbf{{1.5$\uparrow$})}}      \\ 
			\bottomrule
		\end{tabular}
		\label{T1}
	\end{table*}
	
	\begin{table*}[]
		\renewcommand{\arraystretch}{1}
		\setlength{\tabcolsep}{2.mm}
		\caption{Computational efficiency analysis of the models with different settings on the MSRA-TD500 benchmark. ``ED'' and ``Post'' indicates the two specific sub-pipelines (encoder+decoder and post-processing) of TPF in the inference process (details can be referred in Fig.~\ref{V41}). ``Overall gain'' denotes the overall time cost gains brought by ``ED'' and ``Post''. The model is not pre-trained on any public datasets.}
		\centering
		\small
		\begin{tabular}{lccccc}
			\toprule
			\multirow{2}{*}{Methods} & \multirow{2}{*}{Params(M)$\downarrow$} & \multirow{2}{*}{GFLOPs$\downarrow$} & \multicolumn{3}{c}{Time cost(ms)$\downarrow$}                           \\ \cmidrule{4-6} 
			&                            &                         & \multicolumn{1}{c}{ED}  & \multicolumn{1}{c}{Post} & Overall gain \\ \midrule
			baseline                 & 13.0                       & 102.7                   & \multicolumn{1}{c}{21.5} & \multicolumn{1}{c}{8.3}  & --   \\ 
			baseline+REU             & 13.4~(\textbf{{0.4$\uparrow$})}                       & 119.6~(\textbf{{16.9$\uparrow$})}                   & \multicolumn{1}{c}{23.5~(\textbf{{2.0$\uparrow$})}} & \multicolumn{1}{c}{4.1~(\textbf{{4.2$\downarrow$})}}  & \textbf{{2.2$\downarrow$}}  \\ 
			baseline+REU+FPU         & 13.4~(\textbf{\textcolor{gray}{0.0--}})                        & 119.6~(\textbf{\textcolor{gray}{0.0--}})                   & \multicolumn{1}{c}{23.5~(\textbf{\textcolor{gray}{0.0--}}) } & \multicolumn{1}{c}{3.0~(\textbf{{1.1$\downarrow$})}}  & \textbf{{1.1$\downarrow$}}  \\ \bottomrule
		\end{tabular}
		\label{T11}
	\end{table*}
	
	\subsection{Implementation Details}
	The network architecture of the designed TPF is shown in Fig.~\ref{V2}, which consists of the backbone, FPN, and four branches. For comparing our approach with existing methods to show the comprehensive superiorities of TPF in a fair comparison environment, we choose ResNet-18 and ResNet-50 as our backbone. The structure of FPN can be referred to~\cite{lin2017feature} except for the channel of output fusion feature is set to 512. Meanwhile, in the experimental platform and setup, we adopt 4 Nvidia 1080Ti GPUs for pretraining and finetuning our model and 1 Nvidia 1080Ti GPU to evaluate the model performance in the inference process, the choice of the mainstream GPU ensures a fair hardware condition for performance comparisons. Notably, to adapt to the versions of different dependency libraries, we choose a higher version of Pytorch 1.9 to code the TPF framework and the training and testing processes.
	
	\textbf{In the data pre-processing stage}, the image is resized to the specific size proportionally along its short side. Normally, the short sides of the images in MSRA-TD500 and ICDAR2015 datasets are resized to 736. For Total-Text and CTW1500 datasets, they are re-scaled to 640 in this paper. To further explore the model's ability for dealing with different-scaled input, those image sizes will be adjusted to 512 in comparison experiments. Except for re-scaling input images, data augmentation is extra adopted to improve model generalization in the training stage. Concretely, the augmentation consists of the following four steps mainly:(1) random scaling slightly for increasing the text instance scales; (2) random horizontal flipping for providing inverted training samples; (3) random rotating for generating multi-oriented data with diverse angles; (4) random cropping and padding for ensuring a uniform image size in the same batch.
	
	\textbf{In the training stage}, the CNN layers of backbone, FPN, and four branches have to be initialized first. For the backbone, we load the ResNet that is pre-trained on ImageNet~\cite{deng2009imagenet} into our model to initialize the corresponding layers. For the other parts of TPF, we initialize them via normal distribution. The training stage includes the two sub-stages of pre-training and fine-tuning. The proposed TPF is pre-trained on SynthText~\cite{gupta2016synthetic} by 1 epoch and is fine-tuned on other official datasets (such as MSRA-TD500, CTW1500, and so on) by 1200 epochs with a batch size of 16. In the backward process of the training stage, Adam optimizer is chosen to propagate the gradient. The initial learning rate is set as 0.001 and is decayed via the `PolyLr' strategy.
	
	\begin{figure*}
		\centering
		\includegraphics[width=.98\textwidth]{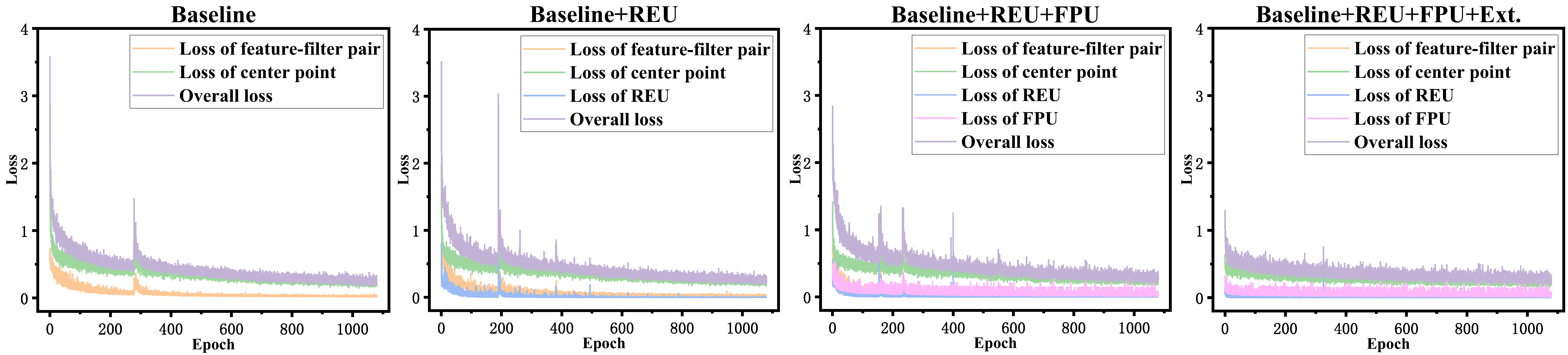}
		\caption{Visualization of the training processes. We show the training details of the models of `baseline', `baseline+REU', `baseline+REU+FPU', and `baseline+REU+FPU+Ext.' from left to right, respectively.}
		\label{V7}
	\end{figure*}

	\subsection{Ablation Study and Hyperparameter Tuning}
	\label{Ablation Study}
	In this section, we demonstrate the effectiveness of REU and FPU for improving detection accuracy and analyze the inference speed gains brought by REU. Meanwhile, we analyze the impacts of different weighted FPU loss (${\cal L}_{fpu}$) (can be referred to~\ref{Loss Function}) and different numbers of center points on model performance. Furthermore, the model's ability for dealing with different-scaled input images is explored. The details of the corresponding experiments are shown next.

	\begin{table}[]
		\renewcommand{\arraystretch}{1}
		\setlength{\tabcolsep}{4.mm}
		\caption{Performance analysis of the models with different weighted FPU losses on the MSRA-TD500 dataset. `$\alpha$' denotes the weight of FPU loss. $\dagger$ means the model is not pre-trained on the SynthText dataset.}
		\centering
		\begin{tabular}{cccc}
			\toprule
			\multicolumn{1}{c}{$\alpha$} & \multicolumn{1}{c}{Presicion} & \multicolumn{1}{c}{Recall} & F-measure \\ 
			\midrule
			\multicolumn{1}{c}{0.1$\dagger$}    & \multicolumn{1}{c}{88.0}      & \multicolumn{1}{c}{80.3}   & 84.0      \\ 
			\multicolumn{1}{c}{0.2$\dagger$}    & \multicolumn{1}{c}{87.8}      & \multicolumn{1}{c}{81.2}   & 84.4      \\ 
			\multicolumn{1}{c}{0.3$\dagger$}    & \multicolumn{1}{c}{88.7}      & \multicolumn{1}{c}{81.1}   & 84.7      \\ 
			\multicolumn{1}{c}{0.4$\dagger$}    & \multicolumn{1}{c}{88.8}      & \multicolumn{1}{c}{81.9}   & 85.2      \\ 
			\multicolumn{1}{c}{0.5$\dagger$}    & \multicolumn{1}{c}{90.3}      & \multicolumn{1}{c}{80.7}   & 85.2      \\ 
			\multicolumn{1}{c}{0.6$\dagger$}    & \multicolumn{1}{c}{90.0}      & \multicolumn{1}{c}{80.7}   & 85.1      \\ 
			\multicolumn{1}{c}{0.7$\dagger$}    & \multicolumn{1}{c}{89.7}      & \multicolumn{1}{c}{81.7}   & {85.5}      \\ 
			\multicolumn{1}{c}{0.8$\dagger$}    & \multicolumn{1}{c}{85.8}      & \multicolumn{1}{c}{83.4}   & 84.6      \\ 
			\multicolumn{1}{c}{0.9$\dagger$}    & \multicolumn{1}{c}{88.4}      & \multicolumn{1}{c}{81.5}   & 84.8      \\ 
			\multicolumn{1}{c}{1.0$\dagger$}      & \multicolumn{1}{c}{89.2}      & \multicolumn{1}{c}{80.5}   & 84.6      \\ 
			\bottomrule
		\end{tabular}
		\label{T2}
	\end{table}
	
\begin{figure*}
		\centering
		\includegraphics[width=.85\textwidth]{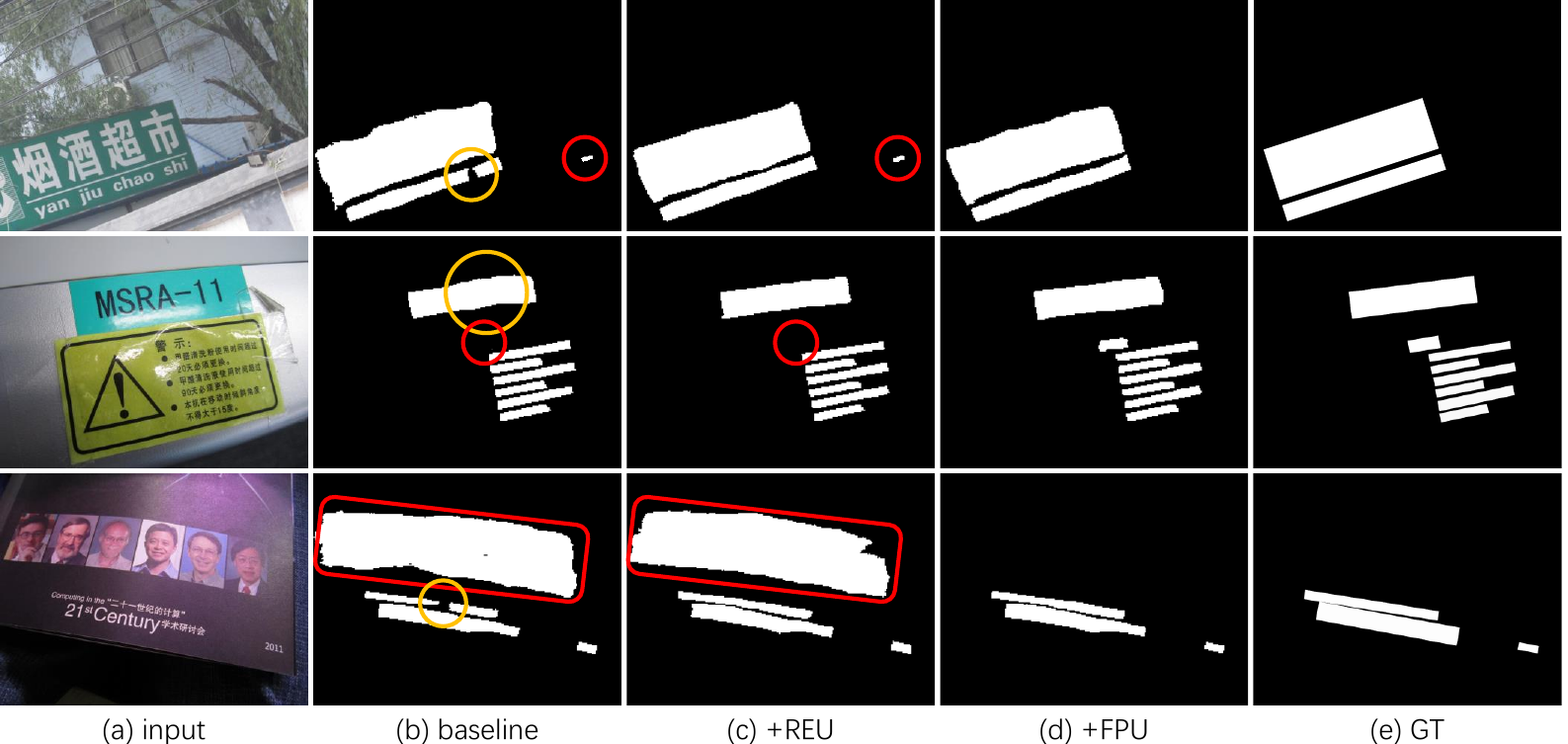}
		\caption{Visualization of the detection results of the models with different settings. The orange circled regions are the high-quality results improved by REU. The red circled regions show the hard positive samples are detected and the false-positive samples are suppressed with the help of strong distinguishment ability to the foreground brought by FPU.}
		\label{V8}
	\end{figure*}

	\textbf{Effectiveness and Efficiency of REU.} As described in Section~\ref{Reinforcement Ensemble Unit}, REU is designed for enhancing feature consistency of the same text while strengthening filter's recognition ability to improve the quality of the detection result. To verify the effectiveness of REU, we compare the experimental results of the models with REU and without it in Table~\ref{T1}~\#1--\#2. It can be found that the enhanced feature consistency and the strengthened filter brought by REU brings 1.8\% improvements in the performance of precision. Meanwhile, we show the training process and improvement by visualizing the predicted binary masks in Fig.~\ref{V7} and Fig.~\ref{V8}, respectively. Concretely, in the orange circled regions of Fig.~\ref{V8}(b) and Fig.~\ref{V8}(c), it is observed that REU helps to recognize the text instance detail more accurately. Except for the precision, REU helps TPF obtain 1.5\% gains in the performance of recall (as shown in Table~\ref{T1}~\#2). We explain the reason in the third row of Fig.~\ref{V8}(b) and Fig.~\ref{V8}(c). The visualization shows that REU can encourage our model to avoid detecting one text instance as multiple parts, which improves the recall rate of detection results effectively. The performance gains in both aspects of precision and recall bring 1.7\% improvements in the comprehensive performance of F-measure.The above experimental results demonstrate the effectiveness of the designed REU.

	Benefiting from the advantages of ensemble operation of the predicted filters (as illustrated in Fig.~\ref{V4} and Algorithm~\ref{algorithm1}), REU can speed up the inference process efficiently. It is verified in Table~\ref{T11}~\#1--\#2. Though the REU branch brings 16.9 GFLOPs to our model in the inference process, it improves the feature consistency and enlarges the filter recognition field, which helps suppress many low-quality detection results. The enhancement of detection results helps save 50.6\% computational costs for the post-processing of TPF. Therefore, the model with REU can bring performance gains in both two aspects of detection accuracy and speed simultaneously.
	
	\textbf{Effectiveness and Efficiency of FPU.} It can be found in Fig.~\ref{V2} and Fig.~\ref{V4} that the proposed TPF first locates text instances according to the predicted center points and then extract feature-filter combined with the point coordinates for detecting texts. To improve the reliability of the center point, FPU is introduced to enhance the model's ability to foreground recognition. As shown in Table~\ref{T1}, FPU brings 2.1\% improvements for the recall rate of detection results, which benefits from the strong distinguishment ability of the foreground from the background brought by FPU. As visualized in the red circled regions of Fig.~\ref{V8}(c) and Fig.~\ref{V8}(d), FPU helps our model recognize hard positive samples. Meanwhile, FPU enhances TPF's ability to suppress false-positive samples (the red boxed regions of Fig.~\ref{V8}(c) and Fig.~\ref{V8}(d)). They are helpful for locating text center points accurately. The above experimental results in Tabla~\ref{T1} and Fig.~\ref{V8} demonstrate the performance gains for predicting center points brought by FPU. Moreover, as we mentioned before, FPU is proposed to help locate text instances accurately and do not participate in the text detection directly. Therefore, it brings no extra computational costs to the inference process (as shown in Table~\ref{T11}~\#2--\#3). Benefiting from the advantages of the improvements of text location accuracy, instances that needed to be processed in post-processing are decreased, which makes FPU further save 37\% computational costs compared with the model with REU only.

	\begin{table}[]
		\renewcommand{\arraystretch}{1}
		\setlength{\tabcolsep}{1.mm}
		\caption{Performance analysis of the models with different numbers of center points on the Total-Text dataset. `N' denotes the center point number. $\dagger$ means the model is not pre-trained on the SynthText dataset.}
		\centering
		\begin{tabular}{cccccc}
			\toprule
			N  & Precision & Recall & F-measure & Post cost (ms) & FPS \\ 
			\midrule
			5$\dagger$   &  84.9         &  83.4      &  84.1         &     5.1      &   40.2  \\ 
			10$\dagger$ &  87.2         &  82.8      &  84.9         &      7.9     &    36.1  \\ 
			15$\dagger$ &  87.2         &  84.1      &  85.6         &      11.9     &   31.5  \\  
			\bottomrule
		\end{tabular}
		\label{T3}
	\end{table}
	
	\textbf{Impacts of Different Weighted FPU Losses.} The experimental results in Table~\ref{T1} and Fig.~\ref{V8} show the strong recognition of the foreground brought by FPU can help our model suppress false-positive samples and detect hard-positive samples, which brings gains for detection performance effectively. To further explore the effectiveness of FPU, we evaluate the performance of the models with different weighted FPU Losses. As shown in Table~\ref{T2}, we tune the FPU loss weight $\alpha$ from 0.2 to 1.0 for testing model performance, respectively. It can be observed that there is a significant gain when $\alpha$ is set as 0.7. Compared with the model without FPU (Table~\ref{T1}~\#2), the experimental results demonstrate the effectiveness of FPU in improving the model's ability to distinguish the foreground. Furthermore, the performance continues to slow-degrading when $\alpha$ is tuned smaller or bigger than 0.7. Based on this conclusion, $\alpha$ is set as 0.7 in the next experiments unless otherwise noted. 
	
	\begin{table}[]
		\renewcommand{\arraystretch}{1}
		\setlength{\tabcolsep}{2.6mm}
		\caption{Performance analysis of the models with different-scaled samples on the MSRA-TD500 dataset. `Scale' denotes the short side size of the testing image. $\ddagger$ means the model is pre-trained on the SynthText dataset.}
		\centering
		\begin{tabular}{ccccc}
			\toprule
			\multicolumn{1}{c}{Scale} & \multicolumn{1}{c}{Precision} & \multicolumn{1}{c}{Recall} & \multicolumn{1}{c}{F-measure} & FPS \\ 
			\midrule
			\multicolumn{1}{c}{512$\ddagger$}   & \multicolumn{1}{c}{91.7}      & \multicolumn{1}{c}{82.9}   & \multicolumn{1}{c}{87.1}      &  63.7   \\ 
			\multicolumn{1}{c}{640$\ddagger$}   & \multicolumn{1}{c}{92.7}      & \multicolumn{1}{c}{82.3}   & \multicolumn{1}{c}{87.2}      &  45.2   \\ 
			\multicolumn{1}{c}{736$\ddagger$}   & \multicolumn{1}{c}{91.2}      & \multicolumn{1}{c}{84.6}   & \multicolumn{1}{c}{87.8}      &  34.5   \\ 
			\bottomrule
		\end{tabular}
		\label{T4}
	\end{table}
	
\begin{figure}
		\centering
		\includegraphics[width=.48\textwidth]{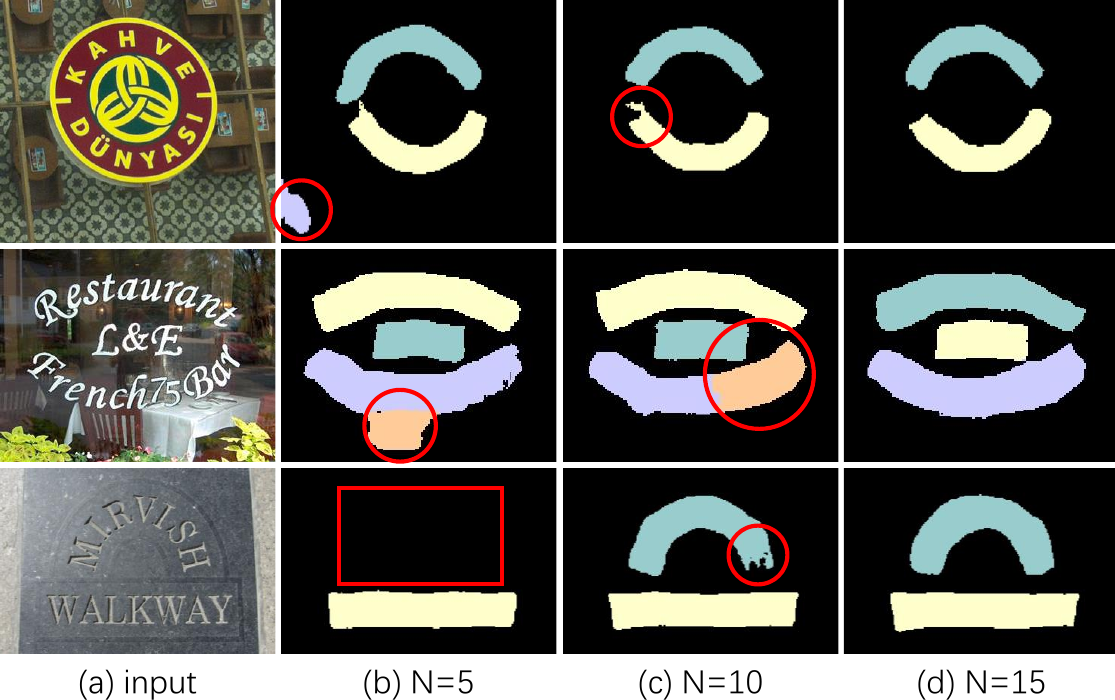}
		\caption{Visualization of the detection results of the models with different numbers of the sampled center points.}
		\label{V10}
	\end{figure}

	\textbf{Impacts of Different Numbers of Center Points.} For the center point prediction header, we sample a specific number of points from each text region as the training label (as shown in Fig.~\ref{V5}). To verify the impacts of different numbers of center points on model performance, we tune the number of the sampled center points from 5 to 10 and 15, respectively. As visualized in Table~\ref{T3}, a larger number of the sampled center points can bring gains for the detection quality. Meanwhile, we show some representative results that are predicted by the models with different center point numbers in Fig~\ref{V10}. It is observed that increasing the number of center point helps distinct the feature difference between the foreground and background (as shown the red circled regions in Fig.~\ref{V10}(b)--(c)). The above experimental results demonstrate that more center points help enhance the recognition of the foreground. However, it leads to our model generating more filters in the inference process, which brings extra computational costs. To achieve a better comprehensive performance between detection accuracy and speed, we choose to sample 5 center points from each text in the experiments of Section~\ref{Comparison with State-of-the-Art Methods}.   
	
	\textbf{Ability for Dealing with Different-Scaled Samples.}
	REU is designed to enhance the filter's recognition capability, enabling the model to effectively handle samples of varying scales and improve its generalization. In this paragraph, we evaluate the model on the MSRA-TD500 dataset using multiple image sizes to demonstrate its scale invariance within a specific range. Specifically, as shown in Table~\ref{T4}, the model's performance remains consistent when the image size is resized from 512 to 640 pixels, achieving the highest accuracy at an image size of 736 pixels. This analysis of both the testing and training processes validates the effectiveness of REU and confirms that TPF performs well across different scales.
	
	\begin{figure*}
		\centering
		\includegraphics[width=.95\textwidth]{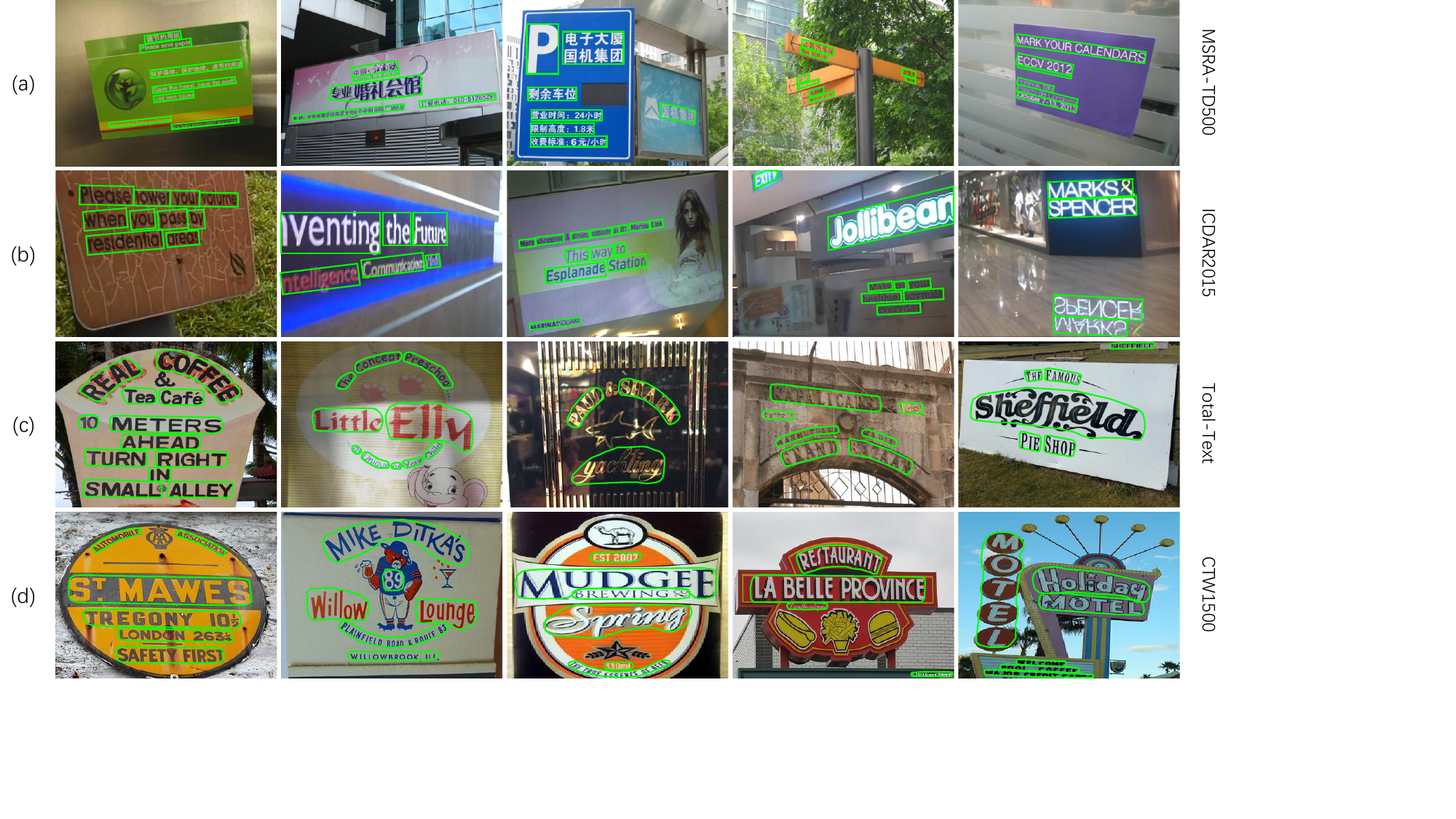}
		\caption{Qualitative detection results of TPF on four public datasets (including the MSRA-TD500, ICDAR2015, Total-Text, and CTW1500 datasets).}
		\label{V12}
	\end{figure*}
	
	\subsection{Comparison with State-of-the-Art Methods}
	\label{Comparison with State-of-the-Art Methods}
	The efficiency of TPF for arbitrary-shaped text detection and the effectiveness of REU and FPU are verified in Section~\ref{Ablation Study}. Meanwhile, we explore the impacts of different weighted FPU losses and different numbers of center points on model performance. Furthermore, the model's ability for dealing with different-scaled samples is verified. The above ablation studies help to understand TPF details and to construct the most efficient framework. In this section, we further to show the superior comprehensive performance of TPF by comparing it with existing state-of-the-art (SOTA) methods under the guidance of the experimental results before. Our model is pre-trained on SynthText (unless otherwise noted) in the next experiments to compare with existing methods.
	
	\begin{table}[]
		\renewcommand{\arraystretch}{1}
		\setlength{\tabcolsep}{1.2mm}
		\caption{Performance comparison with state-of-the-art methods on the MSRA-TD500 benchmark. We highlight the best result of the existing SOTA method and the proposed TPF through ``\textbf{gray background}''. ``Ours-640'' and ``Ours-736'' denote the short sizes of images are resized as 640 and 736 in the inference process, respectively. ResNet18 is employed for multi-scale feature extraction unless stated otherwise.}
		\centering
		\small
		\begin{tabular}{llcccc}
			\toprule
			Methods   & Venue & Precision & Recall & F-measure & FPS  \\ 
			\midrule
			PAN~\cite{wang2019efficient}       & ICCV'19     & 84.4      & 83.8   & 84.1      & 30.2     \\ 
			DB~\cite{liao2020real}        & AAAI'20     & 91.5      & 79.2   & 84.9      & 32.0     \\ 
			MTS-v3~\cite{liao2020mask}        & ECCV'20     & 90.7      & 77.5   & 83.5      &  --    \\ 
			GV~\cite{xu2020gliding}     & TPAMI'20    & 88.8      & 84.3   & 86.5      & 15.0     \\ 
			MCN~\cite{DBLP:journals/ijcv/LiuLG20}        & IJCV'20     & 89.1      & 80.7   & 85.2      &  --    \\ 
			ABPN~\cite{zhang2021adaptive}        & ICCV'21     & 85.4      & 80.7   & 83.0      &  12.7    \\ 
			CM-Net~\cite{DBLP:journals/tip/YangCXYW22}     & TIP'22    & 89.9      & 80.6   & 85.0      & 41.7 \\   
			RFN~\cite{DBLP:journals/tcsv/GuanGLTFWG22}       & TCSVT'22    & 88.4      & 80.0   & 84.0      &  --    \\ 
			PAN++~\cite{DBLP:journals/pami/WangXLLLYLS22}     & TPAMI'22    & 85.3      & 84.0   & 84.7      & 32.5     \\ 
			SPM~\cite{9779460}       & TPAMI'23    & 88.6      & 82.7   & 85.5      &  8.3    \\ 
			DB++~\cite{DBLP:journals/pami/LiaoZWYB23}      & TPAMI'23    & 91.5      & 83.3   & 87.2      & 29.0 \\ 
			MorphText~\cite{xu2023morphtext}      & TMM'23    & 88.5      & 82.7   & 85.5      & -- \\ 
			ABPN~\cite{zhang2023arbitrary}      & TMM'24    & 89.2      & 85.4   & 87.3      & 15.2 \\ 
			\midrule
			\textbf{Ours-640}      & --           & 92.7      & 82.3   & 87.2      &  45.2     \\
			\textbf{Ours-736}      & --           & \cellcolor{gray!30}{91.2}      & \cellcolor{gray!30}{84.6}   & \cellcolor{gray!30}{87.8}      &  \cellcolor{gray!30}{34.5}    \\ 
			\bottomrule
		\end{tabular}
		\label{T5}
	\end{table}
	
	\textbf{Evaluation on MSRA-TD500.} To demonstrate the superiority of TPF for detecting long multi-lingual texts, we compare the detection results with other SOTA works published recently. Particularly, We show the model performance for dealing with different-scaled images to illustrate the superior comprehensive performance of TPF.
	
	Specifically, as we can see from Table~\ref{T5}, our model achieves 87.8\% in F-measure when the input image is resized to 736, which outperforms the SOTA method (BoundTrans~\cite{zhang2023arbitrary}) 0.5\% in F-measure. Particularly, benefiting from the advantages of REU, TPF enjoys a strong ability for dealing with multi-scaled samples, which helps our method achieves comparable detection accuracy with DB++~\cite{DBLP:journals/pami/LiaoZWYB23} when the input is resized to 640 while running 16.2 FPS faster than it. For DB~\cite{liao2020real}, PAN~\cite{wang2019efficient}, and PAN++~\cite{DBLP:journals/pami/WangXLLLYLS22}, TPF outperforms them a lot in both detection accuracy and speed simultaneously, where the important reason is that different from previous whole region-based methods, TPF simulates the operation mode of band-pass filter to detect texts. Therefore, it can separate adhesive texts naturally without extra decoding or post-processing processes. To better show the model's ability to detect long multi-lingual texts, we show some qualitative detection results in Fig.~\ref{V12}(a). The above experimental results verify the effectiveness of TPF for detecting long text lines on the samples of the MSRA-TD500 benchmark.
	
	\begin{table}[]
		\renewcommand{\arraystretch}{1}
		\setlength{\tabcolsep}{1.0mm}
		\caption{Performance comparison with state-of-the-art methods on the ICDAR2015 benchmark. We highlight the best result of the existing SOTA method and the proposed TPF through ``\textbf{gray background}''. ``Ours-736-R50'' means the short sizes of images are resized as 736 and the ResNet50 is adopted as the backbone. ResNet18 is employed for multi-scale feature extraction unless stated otherwise.}
		\centering
		\small
		\begin{tabular}{llccccc}
			\toprule
			Methods   & Venue & Precision & Recall & F-measure & FPS  \\ 
			\midrule
			MSR~\cite{DBLP:conf/ijcai/XueLZ19}       & IJCAI'19    & 86.6      & 78.4   & 82.3      &  4.3    \\ 	
			PAN~\cite{wang2019efficient}       & ICCV'19    & 84.0      & 81.9   & 82.9      &  26.1    \\ 	
			DR-PSS~\cite{DBLP:journals/tcsv/ChengCW20}       & TCSVT'20    & 87.0      & 81.0   & 84.0      &  6.7    \\ 	
			Boundary~\cite{wang2020all}       & AAAI'20    & 82.2      & 88.1   & 85.0      &  --    \\ 	
			FCE-Net\cite{zhu2021fourier}     & CVPR'21    & 85.1      & 84.2   & 84.6      & --     \\ 
			SLN~\cite{DBLP:journals/tcsv/CaiLCDZWY21}       & TCSVT'21    & 88.0      & 83.0   & 85.0      &  11.2    \\
			TextDCT~\cite{9807447}         & TMM'22    & 86.9      & 83.7   & 85.3      &  7.6    \\  
			PATD~\cite{DBLP:journals/tcsv/KeserwaniSLR22}       & TCSVT'22    & 85.9      & 82.7   & 84.3      &  --    \\  
			RFN~\cite{DBLP:journals/tcsv/GuanGLTFWG22}       & TCSVT'22    & 88.4      & 80.0   & 84.0      &  --    \\
			KPN~\cite{zhang2022kernel}     & TNNLS'22    & 84.1      & 83.2   & 83.6      & 12.2     \\ 
			PAN++~\cite{DBLP:journals/pami/WangXLLLYLS22}     & TPAMI'22    & 88.7      & 80.7   & 84.5      & 19.2     \\ 
			SPM~\cite{9779460}     & TPAMI'23    & 88.2      & 83.3   & 85.7      & --     \\ 
			PFText~\cite{wang2023region}     & TMM'23    & 89.6     & 82.4   & 85.9      & --     \\ 
			\midrule
			\textbf{Ours-736}      & --           &    87.8    &   84.0   &   85.9    &  28.7    \\
			\textbf{Ours-736-R50}      & --           &    \cellcolor{gray!30}{91.1}    &   \cellcolor{gray!30}{84.4}   &   \cellcolor{gray!30}{87.6}    &  \cellcolor{gray!30}{11.3}    \\
			\bottomrule
		\end{tabular}
		\label{T6}
	\end{table}
	
	\textbf{Evaluation on ICDAR2015.} We have demonstrated the TPF's strong ability to detect line-level multi-oriented texts before. In this section, we testing our method on ICDAR2015 benchmark to further verify the superiority for detecting word-level multi-oriented texts from the complex background.
	
	In Table~\ref{T6}, existing SOTA methods, such as PFText~\cite{wang2023region}, PAN++~\cite{DBLP:journals/pami/WangXLLLYLS22}, and SLN~\cite{DBLP:journals/tcsv/CaiLCDZWY21}, can achieve 85.9\%, 84.5\%, and 85.0\% in F-measure respectively. Specifically, PAN++ and SLN can run 19.2 and 11.2 FPS in the inference process. They enjoy a comparable comprehensive performance. For the proposed TPF, FPU encourages it to recognize the center points more accurately, which helps detect text instances from the background with plenty of noise. Based on the above advantages, TPF performs better than previous works for the detection task on the ICDAR2015 dataset. Concretely, it outperforms the SOTA method SLN 0.9\% in detection accuracy and 17.5 FPS in detection speed. Though PAN~\cite{wang2019efficient} can run 26.1 FPS, our method outperforms it 3.0\% in F-measure. Furthermore, some results are shown in Fig.~\ref{V12}(b), where text instances can be recognized clearly even though there are many interference regions in the background. Combining the detection results illustrated in Table~\ref{T6} and visualized in Fig.~\ref{V12}(b), we verify the remarkable performance of TPF to detect word-level multi-oriented texts successfully.
	
	\textbf{Evaluation on Total-Text and CTW1500.} Different from the datasets before, Total-Text consists of plenty of word-level curved text instances, which are used for testifying the model's ability to detect texts with irregular shapes separately. Meanwhile, except for the Total-Text dataset, we further testing TPF on the CTW1500 benchmark and compare it with previous methods to show the superior performance for recognizing ribbon-like curved texts and ensuring the integrity of them.
	
	Considering the different features between Transformer~\cite{vaswani2017attention} and CNN, we show some representative methods of the two structures in Table~\ref{T7} top and bottom sections, respectively. As the best CNN-based method for arbitrary-shaped text detection published recently, MorphText~\cite{xu2023morphtext} can achieve 86.9\% and 86.0\% in F-measure on Total-Text and CTW1500 datasets. Though the above method surpasses the other SOTA methods a lot in F-measure, the slow running speed limits further performance improvement. Furthermore, in the aspect of detection speed, DB++~\cite{DBLP:journals/pami/LiaoZWYB23} and PAN++ run faster almost 4 times than existing related works~\cite{9807447}. Though DB++ performs well in both detection accuracy and speed on the Total-Text and CTW1500 datasets, our method still can achieve a better comprehensive performance than it. As we introduced before, TPF segments the whole regions of texts directly, which avoids the limitation of destroyed semantic integrity and confused feature differences between the foreground and background that exists in shrink-mask-based methods. Meanwhile, it can separate adhesive texts without complex decoding or post-processing processes. The above advantages bring gains to the comprehensive performance of TPF. Specifically, our model outperforms DB++ 2.3\% and 2.0\% in F-measure on Total-Text and CTW1500 benchmarks respectively and runs 10 FPS faster than it at least. Moreover, for the accuracy prior method (ContourNet), TPF still can surpass it 0.2\% in F-measure while running almost 10 times faster than it. Compared to current SOTA transformer-based methods (DPText~\cite{ye2023dptext} and DeepSolo++~\cite{ye2023deepsolo++}), our method performs worse in terms of F-measure but achieves faster inference speed. The differences in accuracy and efficiency among these methods primarily stem from the distinctions between transformer and CNN architectures. In addition to performance differences, their hardware resource dependencies vary significantly. Transformer-based methods require GPUs with large memory (e.g., 4090 GPU or A100 GPU), which may be inaccessible to many researchers. In contrast, CNN-based detectors can typically be trained and evaluated effectively on more accessible hardware like the 1080Ti GPU. The CNN-based methods allow researchers without enough hardware resources to promote the development of efficient scene text methods in the aspect of framework design.

	\begin{table}[]
		\renewcommand{\arraystretch}{1}
		\setlength{\tabcolsep}{1.2mm}
		\caption{Comparisons on the Total-Text and CTW1500. We highlight the best result of different methods through ``\textbf{gray background}''. ``Ours-640-R50'' means the short sizes of images are resized as 640 and the ResNet50 is adopted as the backbone. ResNet18 is employed by default for multi-scale feature extraction. Methods in the bottom section do not adopt a transformer structure.}
		\centering
		\small
		\begin{tabular}{lcccccccc}
			\toprule
			\multirow{2}{*}{Methods} & \multicolumn{4}{c}{Total-Text}                        &\multicolumn{4}{c}{CTW1500}          \\\cmidrule{2-9}
			&                 \multicolumn{1}{c}{P} & \multicolumn{1}{c}{R} & \multicolumn{1}{c}{F} & FPS  & \multicolumn{1}{c}{P} & \multicolumn{1}{c}{R} & \multicolumn{1}{c}{F} & FPS  \\ 
			\midrule
			STKM~\cite{wan2021self}                  &  \multicolumn{1}{c}{86.3}      & \multicolumn{1}{c}{78.4}   & \multicolumn{1}{c}{82.2}      & --   & \multicolumn{1}{c}{85.1}      & \multicolumn{1}{c}{78.2}   & \multicolumn{1}{c}{81.5}      & --   \\ 
			ABPN~\cite{zhang2023arbitrary}                                      & \multicolumn{1}{c}{89.9}      & \multicolumn{1}{c}{85.3}   & \multicolumn{1}{c}{87.5}      & 12.0 & \multicolumn{1}{c}{88.1}      & \multicolumn{1}{c}{81.1}   & \multicolumn{1}{c}{84.5}      & 14.7 \\ 
			TextPMs~\cite{zhang2022arbitrary}                                      & \multicolumn{1}{c}{89.8}      & \multicolumn{1}{c}{87.8}   & \multicolumn{1}{c}{87.2}      & 6.8 & \multicolumn{1}{c}{87.6}      & \multicolumn{1}{c}{80.8}   & \multicolumn{1}{c}{84.1}      & 9.0 \\ 
			DPText~\cite{ye2023dptext}                                      & \multicolumn{1}{c}{\cellcolor{gray!30}{91.8}}      & \multicolumn{1}{c}{\cellcolor{gray!30}{86.4}}   & \multicolumn{1}{c}{\cellcolor{gray!30}{89.0}}      & \cellcolor{gray!30}{--} & \multicolumn{1}{c}{91.7}      & \multicolumn{1}{c}{86.2}   & \multicolumn{1}{c}{88.8}      & -- \\ 
			DeepSolo++~\cite{ye2023deepsolo++}                                      & \multicolumn{1}{c}{93.9}      & \multicolumn{1}{c}{82.1}   & \multicolumn{1}{c}{87.6}      & -- & \multicolumn{1}{c}{\cellcolor{gray!30}{92.5}}      & \multicolumn{1}{c}{\cellcolor{gray!30}{86.3}}   & \multicolumn{1}{c}{\cellcolor{gray!30}{89.3}}      & \cellcolor{gray!30}{--} \\ 
			\midrule
			OPMP~\cite{zhang2020opmp}                  &  \multicolumn{1}{c}{87.6}      & \multicolumn{1}{c}{82.7}   & \multicolumn{1}{c}{85.1}      & 1.4 
			& \multicolumn{1}{c}{85.1}      & \multicolumn{1}{c}{80.8}   & \multicolumn{1}{c}{82.9} 
			& 1.4 \\ 
			FCE-Net~\cite{zhu2021fourier}                               & \multicolumn{1}{c}{87.4}      & \multicolumn{1}{c}{79.8}   & \multicolumn{1}{c}{83.4}      & --   & \multicolumn{1}{c}{85.7}      & \multicolumn{1}{c}{80.7}   & \multicolumn{1}{c}{83.1}      & --   \\ 
			RDSS~\cite{DBLP:journals/ijcv/FengYZHL21}                                   & \multicolumn{1}{c}{87.1}      & \multicolumn{1}{c}{80.3}   & \multicolumn{1}{c}{83.5}      & --   & \multicolumn{1}{c}{87.3}      & \multicolumn{1}{c}{81.8}   & \multicolumn{1}{c}{84.5}      & --   \\  
			NASK~\cite{DBLP:journals/tcsv/CaoZYZ22}                                     & \multicolumn{1}{c}{91.1}      & \multicolumn{1}{c}{52.5}   & \multicolumn{1}{c}{66.6}      & --   & \multicolumn{1}{c}{83.4}      & \multicolumn{1}{c}{80.1}   & \multicolumn{1}{c}{81.7}      & 12.1 \\
			ABS-Net~\cite{DBLP:journals/tcsv/NandanwarSRLPAL22}                  &  \multicolumn{1}{c}{85.6}      & \multicolumn{1}{c}{83.2}   & \multicolumn{1}{c}{84.4}      & 8.4  & \multicolumn{1}{c}{92.7}      & \multicolumn{1}{c}{74.4}   & \multicolumn{1}{c}{82.4}      & --   \\
			TextDCT~\cite{9807447}                  &  \multicolumn{1}{c}{85.8}      & \multicolumn{1}{c}{80.5}   & \multicolumn{1}{c}{83.0}      & 15.2  & \multicolumn{1}{c}{84.7}      & \multicolumn{1}{c}{81.5}   & \multicolumn{1}{c}{83.1}      & 17.3   \\ 
			KPN~\cite{zhang2022kernel}                    &  \multicolumn{1}{c}{88.0}      & \multicolumn{1}{c}{82.3}   & \multicolumn{1}{c}{85.1}      & 22.7 & \multicolumn{1}{c}{84.0}      & \multicolumn{1}{c}{82.9}   & \multicolumn{1}{c}{83.4}      & 24.3 \\ 
			PAN++~\cite{DBLP:journals/pami/WangXLLLYLS22}                                     & \multicolumn{1}{c}{89.9}      & \multicolumn{1}{c}{81.0}   & \multicolumn{1}{c}{85.3}      & 38.3 & \multicolumn{1}{c}{87.1}      & \multicolumn{1}{c}{81.1}   & \multicolumn{1}{c}{84.0}      & 36.0 \\ 
			DB++~\cite{DBLP:journals/pami/LiaoZWYB23}                                      & \multicolumn{1}{c}{87.4}      & \multicolumn{1}{c}{79.6}   & \multicolumn{1}{c}{83.3}      & 48.0 & \multicolumn{1}{c}{84.3}      & \multicolumn{1}{c}{81.0}   & \multicolumn{1}{c}{82.6}      & 49.0 \\ 
			MorphText~\cite{xu2023morphtext}                                      & \multicolumn{1}{c}{88.4}      & \multicolumn{1}{c}{85.5}   & \multicolumn{1}{c}{86.9}      & -- & \multicolumn{1}{c}{89.0}      & \multicolumn{1}{c}{83.2}   & \multicolumn{1}{c}{86.0}      & -- \\ 
			\textbf{Ours-512}                                   & \multicolumn{1}{c}{86.4}          & \multicolumn{1}{c}{82.5}       & \multicolumn{1}{c}{84.4}          &  {{58.9}}    & \multicolumn{1}{c}{85.0}          & \multicolumn{1}{c}{83.1}       & \multicolumn{1}{c}{84.0}          &  {{60.6}}    \\ 
			\textbf{Ours-640}                                             & \multicolumn{1}{c}{87.1}          & \multicolumn{1}{c}{84.3}       & \multicolumn{1}{c}{{{85.7}}}          &   34.0   & \multicolumn{1}{c}{86.8}          & \multicolumn{1}{c}{82.5}       & \multicolumn{1}{c}{{{84.6}}}          &   41.2   \\ 
			\textbf{Ours-640-R50}                                             & \multicolumn{1}{c}{\cellcolor{gray!30}{88.5}}          & \multicolumn{1}{c}{\cellcolor{gray!30}{85.3}}       & \multicolumn{1}{c}{\cellcolor{gray!30}{86.9}}          &   \cellcolor{gray!30}{17.4}   & \multicolumn{1}{c}{\cellcolor{gray!30}{87.9}}          & \multicolumn{1}{c}{\cellcolor{gray!30}{84.5}}       & \multicolumn{1}{c}{\cellcolor{gray!30}{86.2}}          &   \cellcolor{gray!30}{19.1}   \\ 
			\bottomrule
		\end{tabular}
		\label{T7}
	\end{table}
	
	\subsection{Analysis of Model Limitations}
\begin{figure}
		\centering
		\includegraphics[width=.48\textwidth]{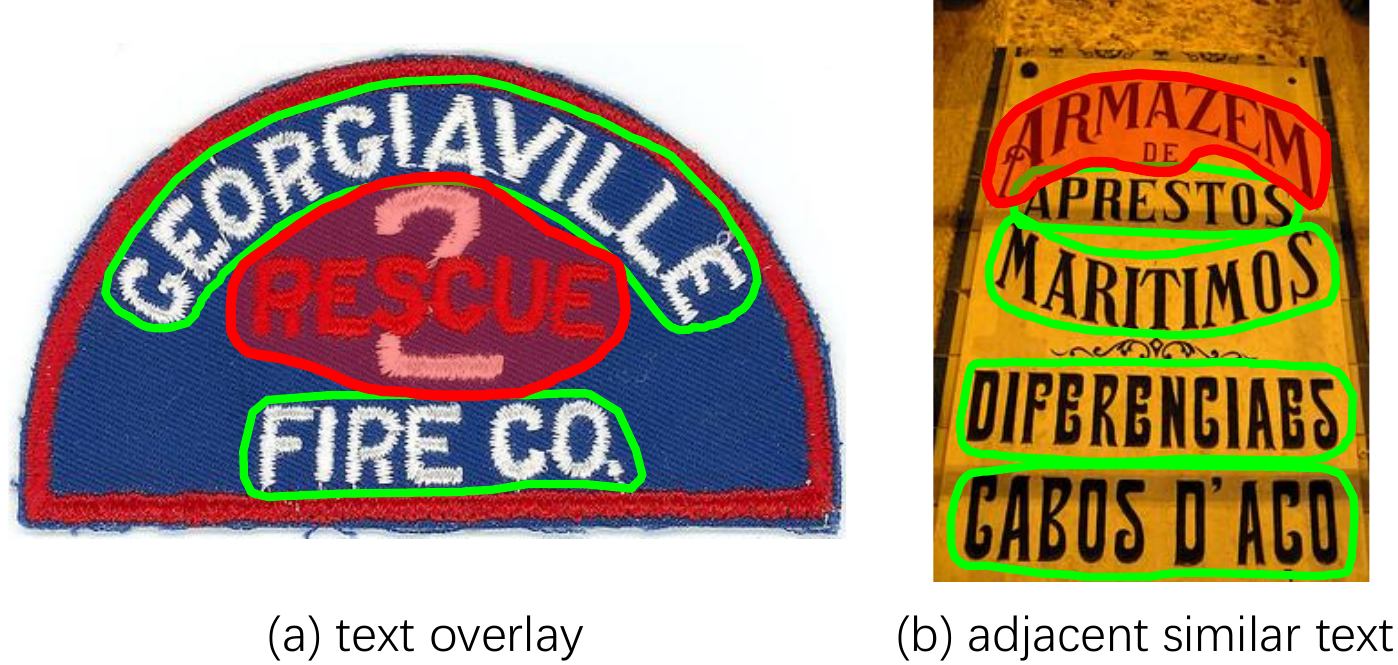}
		\caption{Visualization of challenging samples. The green contours are the correct predicted results and the red ones are the failure cases.}
		\label{V13}
	\end{figure}

	We have explored the importance of different factors affecting model performance and have demonstrated the superior performance of the proposed TPF on multiple public datasets before. Further, to help understand TPF comprehensively, we discuss the weaknesses of TPF in this section, illustrating scenarios that our model finds hard to handle. As shown in Fig.~\ref{V13}, there are two types of representative challenging samples for our method. Firstly, text overlay (referred to Fig.~\ref{V13}(a)) is a classic challenge to scene text detection. Current methods distinguish texts from the background via the recognition of visual features. These methods lack the semantic analysis from high-level for each latent text region, which is an intrinsic deficiency. Similarly, our TPF is hard to handle this case either. Besides, as we described in Section~\ref{Methodology}, REU merges multiple filters that belong to the same text to a strengthen filter to improve the accuracy of predicted results. However, during the strengthening process, some filters with similar visual features that belong to different texts may be merged together by REU. This causes the strengthened filter to allow multiple texts to pass instead of just one, leading to an overdetection problem (referred to Fig.~\ref{V13}(b)). Therefore, there are still challenges to be considered when applying existing text detection algorithms in real-world scenarios, and vision-language models will be introduced in this field to help alleviate residual problems.
	
	\section{Conclusion}
	\label{Conclusion}
	In this paper, we are inspired by electronics and signal processing to design a text detector, namely Text-Pass Filter (TPF), to simulate the band-pass filter for detecting arbitrary-shaped texts efficiently even though texts are close to each other. It ensures the text's semantic integrity and avoids confusion about the feature difference between the foreground and background to enhance the model's text recognition ability. Meanwhile, TPF can separate adhesive texts according to their unique features without complex decoding and post-processing processes to make it possible for real-time text detection. Furthermore, a Reinforcement Ensemble Unit (REU) is designed to enhance the text feature consistency and to enlarge the filter's recognition field for improving the precision performance. In the end, a Foreground Prior Unit (FPU) is introduced to guide our model to distinguish texts from the background for bringing gains in the recall rate of detection results. Experiments on multiple public datasets demonstrate the superior efficiency of TPF compared with existing methods and the effectiveness of REU and FPU.

\bibliographystyle{IEEEtran}
\bibliography{egbib}

\newpage

\vfill

\end{document}